\documentclass{article}

\usepackage{PRIMEarxiv}

\usepackage[utf8]{inputenc} 
\usepackage[T1]{fontenc}    
\usepackage{hyperref}       
\usepackage{url}            
\usepackage{booktabs}       
\usepackage{nicefrac}       
\usepackage{microtype}      
\usepackage{lipsum}
\usepackage{fancyhdr}       
\usepackage{graphicx}       
\graphicspath{{media/}}     
\usepackage{amssymb}
\usepackage{lineno,hyperref}
\usepackage{amsmath, amsfonts}
\usepackage{amsthm}
\usepackage{bm}
\usepackage{algorithm}
\usepackage{algorithmicx}
\usepackage{algpseudocode}
\usepackage{multirow}
\usepackage{booktabs}       
\usepackage{nicefrac}       
\usepackage{microtype}      
\usepackage{xcolor}         
\usepackage{algpseudocode}
\usepackage{colortbl}
\usepackage{makecell}

\title{Confounder Balancing in Adversarial Domain Adaptation for Pre-Trained Large Models Fine-Tuning}

\author{
  Shuoran Jiang, Qingcai Chen, Xiangping Wu \\
  Haibin Institute of Technology, ShenZhen \\
  ShenZhen\\
   \And
  Yang Xiang, Youchen Pan \\
  Peng Cheng Laboratory \\
  ShenZhen\\
}

\begin{document}
\maketitle

\begin{abstract}
The excellent generalization, contextual learning, and emergence abilities in the pre-trained large models (PLMs) handle specific tasks without direct training data,
making them the better foundation models in the adversarial domain adaptation (ADA) methods to transfer knowledge learned from the source domain to target domains.
However,
existing ADA methods fail to account for the confounder properly,
which is the root cause of the source data distribution that differs from the target domains.
This study proposes an adversarial domain adaptation with confounder balancing for PLMs fine-tuning (ADA-CBF).
The ADA-CBF includes a PLM as the foundation model for a feature extractor,
a domain classifier and a confounder classifier,
and they are jointly trained with an adversarial loss.
This loss is designed to improve the domain-invariant representation learning by diluting the discrimination in the domain classifier.
At the same time,
the adversarial loss also balances the confounder distribution among source and unmeasured domains in training.
Compared to existing ADA methods, ADA-CBF can correctly identify confounders in domain-invariant features,
thereby eliminating the confounder biases in the extracted features from PLMs.
The confounder classifier in ADA-CBF is designed as a plug-and-play and can be applied in the confounder measurable,
unmeasurable,
or partially measurable environments.
Empirical results on natural language processing and computer vision downstream tasks show that ADA-CBF outperforms the newest GPT-4, LLaMA2, ViT and ADA methods.
The source code is released \footnote{https://github.com/MathIsAll/CadaFT.git}.
\end{abstract}


\section{Introduction}
Fine-tuning pre-trained large models (PLMs) for downstream tasks has become a unified learning paradigm for natural language processing (NLP) \cite{he2021towards, ding2023parameter}, computer vision (CV) \cite{sohn2023visual}, multi-modal learning (MM) \cite{sung2022vl}, and other related fields.
The PLMs use unsupervised, self-supervised, reinforcement learning from human feedback (RLHF) and other methods to transfer knowledge from a large scale of the pre-training corpus \cite{korbak2023pretraining},
such as images and texts,
into the parameters of the neural network.
When the PLMs are applied to specific downstream tasks,
they can be quickly adapted via fine-tuning a small amount of data.
However,
the large scale of the pre-training corpus still suffers from the domain imbalance problem \cite{kim2022broad}.
For example,
some areas cannot be covered in the pre-training corpus, or only a small part of the knowledge is involved.
Therefore,
out-of-distribution (OOD) generalization and domain adaptation remain long-standing challenges for PLMs \cite{wang2023robustness}.

\begin{figure}[htbp]
    \centering
    \includegraphics[width=10cm]{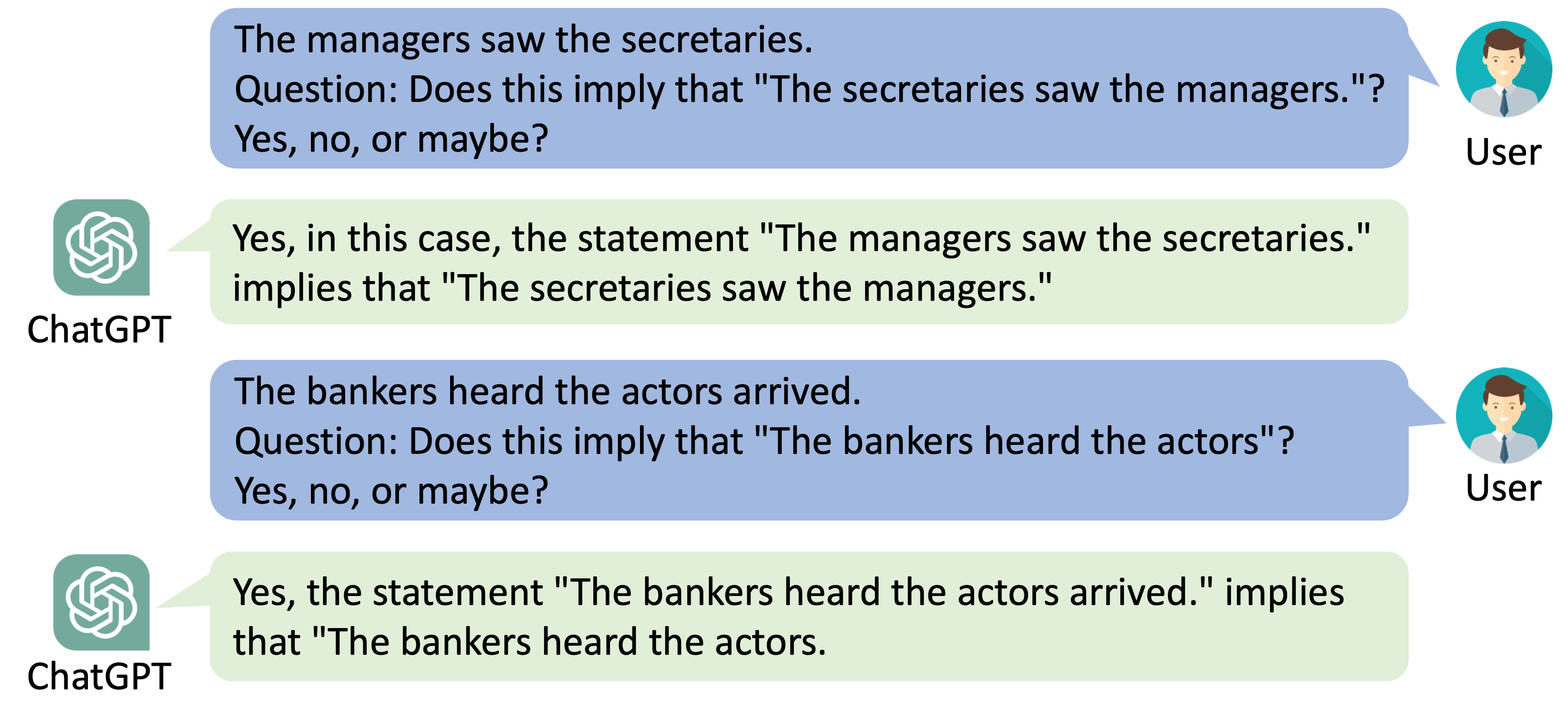}
    \caption{Examples that GPT-3.5-turbo incorrectly predicts that premise and hypothesis with high word overlaps have entailment relationship.}
    \label{fig:llmdemo}
\end{figure}

\begin{figure}[htbp]
    \centering
    \includegraphics[width=10cm]{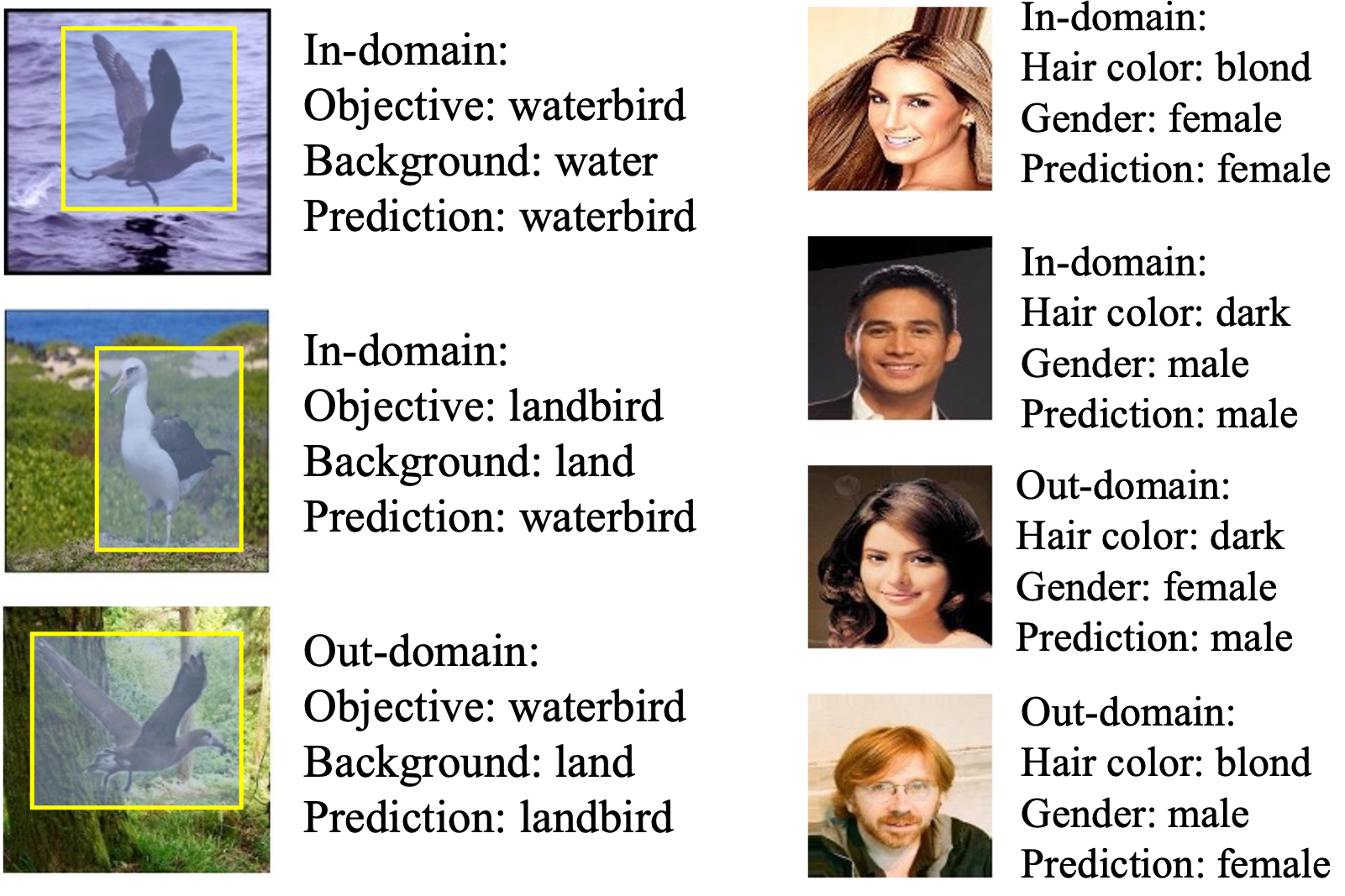}
    \caption{Examples that ViT incorrectly recognizes objects in images depending on the confounders of background in waterbird recognition and hair color in gender prediction.}
    \label{fig:pvmtest}
\end{figure}

The OOD generalization and domain adaptation problems widely exist in real-world applications where data is collected from multiple sources with different characteristics and distributions \cite{sarker2021machine}.
For example,
in general expressions,
two sentences with high word overlaps usually have an entailment relationship in data sources,
like weblogs, news, and online movie reviews.
A language model trained on these corpus will incorrectly respond that two sentences with high lexical overlaps are entailment with each other.
Figure \ref{fig:llmdemo} exhibits two examples to demonstrate this problem in GPT-3.5-turbo \cite{clavie2023large}.
Figure \ref{fig:pvmtest} considers an object recognition task\cite{sagawa2019distributionally} where a vision model is trained to predict whether the bird in pictures is a water bird or a land bird.
Suppose most waterbird pictures in the training data have water backgrounds and most OOD examples have land backgrounds.
In that case,
the vision model may learn to rely on the water background for prediction instead of the cause-effect relationships between object features and output targets.
As a result,
this model can not generalize well to target domains with different backgrounds \cite{ye2021adversarial}.

\begin{figure}[htbp]
    \centering
    \includegraphics[width=10cm]{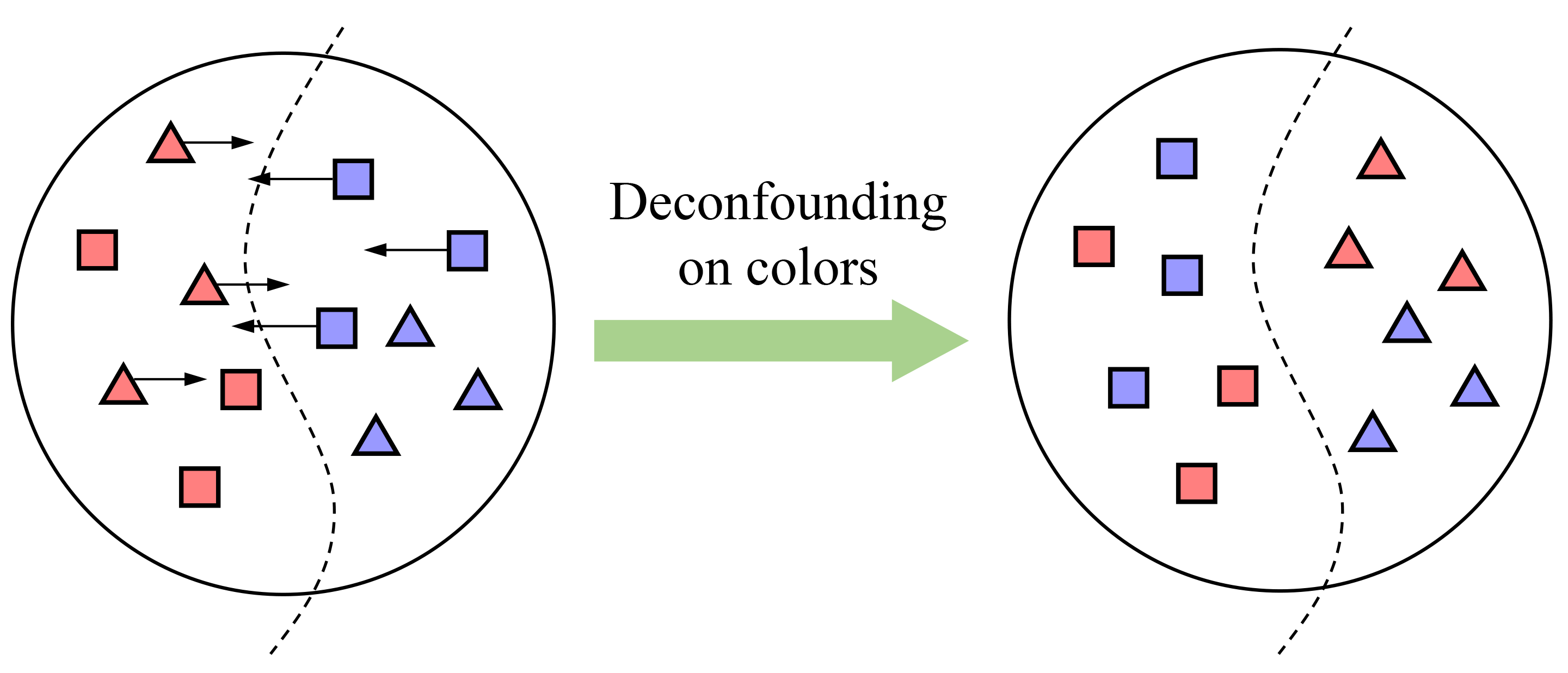}
    \caption{An example of coupled confounder, where the red is observed and blue is unobserved.}
    \label{entangle}
\end{figure}

Large language models (LLMs),
such as GPT-4, LLaMA-2,
and pre-trained vision models (PVMs),
such as ViT \cite{dosovitskiy2020image} and Swin-Transformer \cite{liu2021swin},
have benefited from a large-scale pre-training corpus.
As these data are not specific to a particular domain or task,
these models can learn invariant features and generalize well from one domain to another in downstream tasks \cite{choi2022c2l}.
As a result,
the LLMs and PVMs outperformed previous deep learning models trained from scratch in various fields.
In addition,
the newest LLMs also show powerful in-context learning (ICL) capability \cite{anil2022exploring}.
ICL is a paradigm that allows language models to learn tasks given only a few examples in the form of demonstration \cite{dong2022survey}.
Essentially,
it estimates the likelihood of the potential answer conditioned on the demonstration by using a well-trained language model.
Providing relevant examples may help the model better understand the semantics of the input data,
and maintain certain performance in the face of distribution shifts.
However,
the input length of LLMs limits the number of ICL examples \cite{wang2023augmenting},
and limits fully tap into ICL capability to improve their domain adaptation \cite{wang2023augmenting}.
Taking the PVM as the foundation model,
balanced representation learning (BRL) methods propose a generalization bound to mitigate the bias in the feature space between the source and target domains \cite{shalit2017estimating}.
The underlying idea is to balance the feature spaces of the source and target domains and maximize the margin of source domain feature space \cite{yao2021survey}.
However,
as shown in Figure \ref{entangle},
the explicit balance for one confounding variable (called confounder hereinafter) may unintentionally aggravate bias amplification on other unobserved ones \cite{kang2020exploring}.
Moreover,
BRL methods require exact annatations for confounders,
which are more challenging to obtain in non-quantitative data.
The adversarial domain adaptation (ADA) is a branch of transfer learning,
which utilizes knowledge from a source domain to enhance the performance of target domains \cite{kamath2019deep}.
Existing ADA methods mainly adopt three approaches:
(\romannumeral1) semi-supervised learning with pseudo labeling \cite{zhou2022active},
(\romannumeral2) selecting highly correlated source domains with high similarity \cite{zuo2021attention},
and (\romannumeral3) contra-distinguishing source and target domains to align domain representations \cite{balgi2021contradistinguisher}.
However,
these methods only perform domain-level alignment but do not properly account for the confounders in the feature space \cite{magliacane2018domain, zhang2021deep}.
A confounder relates both the input features and the output predictions in the training dataset,
but is not part of the causal mechanism that generates the predictions \cite{hu2022improving, teshima2020few}.
If confounders are not properly accounted for,
it will degrade the generalization of the model learned from the source domain to the target domain (also known as out-of-distribution (OOD) data) \cite{hsu2020progressive, shen2020stable}.
It could happen if confounders such as the shooting location are related to the object of the pictures but also differ between the source and target domains.
Even though ADA methods can learn domain-invariant features from input data,
they do not balance the confounders to build more reliable predictions.
Methods,
such as stratified sampling, causal inference, and counterfactual reasoning,
can properly account for confounders \cite{pearl2009causal}.
However,
confounders are more challenging to determine as they are not what you see is what you want \cite{raff2019step}
for unstructured data such as text, images, and audio.
For example,
the writer's tone, writing style, and cultural background are most likely the confounders in the text and are not directly measurable or quantifiable.
In addition,
the lighting condition, camera angle, and object placement are also the confounders in images,
which may not be easily separable from the object of interest \cite{friedrich2023fair}.
Moreover,
the more complicated environments are that confounders may interact with each other.

This study proposes a framework of \textbf{c}onfounder balancing in \textbf{a}dversarial \textbf{d}omain \textbf{a}daptation for PLMs \textbf{f}ine-\textbf{t}uning (CadaFT).
Unlike BRL and ADA methods focusing solely on domain-invariant feature learning,
CadaFT properly controls all observed confounders in feature extraction.
which can further help the model to predict based on reliable features.
The proposed CadaFT framework includes the feature extractor, domain classifier, and confounder classifier,
and involves 6 training steps:
(\romannumeral1) Collecting labeled data from the source domain and unlabeled data from the target domain.
(\romannumeral2) Annotating the confounders in both the source domain data and few-shot target domain data.
(\romannumeral3) Extracting feature as the latent representation learning from foundation model, predicting the confounders via a confounder classifier, predictiong the domain from a domain classifier.
(\romannumeral4) Jointly train the feature extractor, confounder classifier and domain classifier with the adversarial loss function.
This training objective is designed to lessen the domain classification accuracy for domain-invariant features,
at the same time to maximize task prediction and confounder classification accuracy,
(\romannumeral5) Fine-tuning the foundation model on the unlabeled data from the source domain to adapt the learned model to the target domain.
In this way,
the bias caused by confounders is removed in the feature extractor when the adversarial loss converges.
Empirical results on downstream NLP and CV tasks demonstrate that CadaFT improves robustness on spurious correlations and achieves new state-of-the-art (SOTA) OOD generalization and domain adaptation.

The main contributions of this paper are listed as follows:
\begin{itemize}
    \item This study proposed a framework of confounder controlling in adversarial domain adaptation (CadaFT) to extract domain-invariant and confounder-irrelevant features from input data.
    \item The proposed CadaFT is a plug-and-play framework for confounder observed or unobserved and or partially observed environments. Furthermore, CadaFT can easily adapt the newest LLMs and PVMs as the feature extractor network. 
    \item The empirical results demonstrated that CadaFT significantly improves the robustness of spurious correlations and outperforms the newest adversarial domain adaptation methods on both NLP and CV tasks.
\end{itemize}

\section{Problem Formulation}

\subsection{Background}

In causal inference,
a confounder is a third-party variable that may be associated with both the dependent and independent variables \cite{pearl2009causality}.
When confounders are not controlled,
it can bias the observed relationships,
thereby the machine learning models cannot make decisions from the relationships of cause and effect \cite{morgan2015counterfactuals}.

\subsection{Definition}

A confounder $C$ (also confounding variable, confounding factor, extraneous determinant or lurking variable),
when it is not considered or controlled,
may lead model to misunderstand the reasons for the relationship between independent variable $X$ and some dependent variable $Y$.
Confounder can give the wrong impression that confounder $C$ is the cause of the dependent variable $Y$,
when in reality,
the independent variable $X$ is the real cause.

If $C$ is not properly accounted for in an analysis, it can introduce bias in the estimation of the causal effect of $T$ on $Y$.

\subsection{Implications}

Failure to account for confounders can lead to:

\begin{itemize}
    \item \textbf{Spurious Relationships:} An observed relationship between $T$ and $Y$ might be due to the influence of $C$ rather than a direct causal effect.
    \item \textbf{Biased Estimations:} The magnitude or direction of the causal effect can be distorted.
    \item \textbf{Misleading Conclusions:} Inferences about causality can be incorrect, potentially leading to wrong decisions or policies.
\end{itemize}

\subsection{Examples}

Consider a CV downstream task of objective recognition,
as shown in Figure \ref{fig:pvmtest},
that evaluates the effect of image $X$ on bird prediction $Y$.
Background $C$ can influence both the likelihood of image and bird prediction and acts as a confounder.
If $C$ is not controlled for,
the observed relationship between image and bird prediction might be biased due to the effects of Background.

Many downstream NLP tasks,
like the natural language inference (NLI) task,
have demonstrated that most language models are sensitive to the shortcuts in the dataset \cite{rajaee2022looking}.
Take the NLI task in Figure \ref{fig:llmdemo} as an example,
there is a confounder $C$ of high word overlap between the entailment label $Y$ and the context $X$ of the premise and hypothesis \cite{mccoy2020right}.
If a language model is trained on the corpus with general representations,
it may depend on the lexical overlaps to predict whether the given premise and hypothesis are entailed.
Moreover,
realistic data usually have diverse and independent confounders.
As shown in Figure \ref{entangle},
the discrepancy minimization on one confounder in traditional balanced representation learning methods can increase the bias on other unobserved ones.

\section{Method}

This section proposed the CadaFT framework,
it assumes that each instance $\mathcal{I} = \left( \bm{x}, \bm{y}, s, \bm{t} \right)$ includes observed feature $\bm{x}\in \mathbb{R}^d$,
prediction target $\bm{y} \in \mathbb{R}^K$,
domain indicator $s \sim \lbrace 0, \cdots, S \rbrace$,
and all measurable confounder set $\bm{t} = \lbrace t_1, \cdots, t_T \rbrace$.
Sometimes,
the confounder set $\bm{t}=\emptyset$ can be an empty set as the confounder in some environments is challenging to annotate.

\begin{figure}[htbp]
    \centering
    \includegraphics[width=14cm]{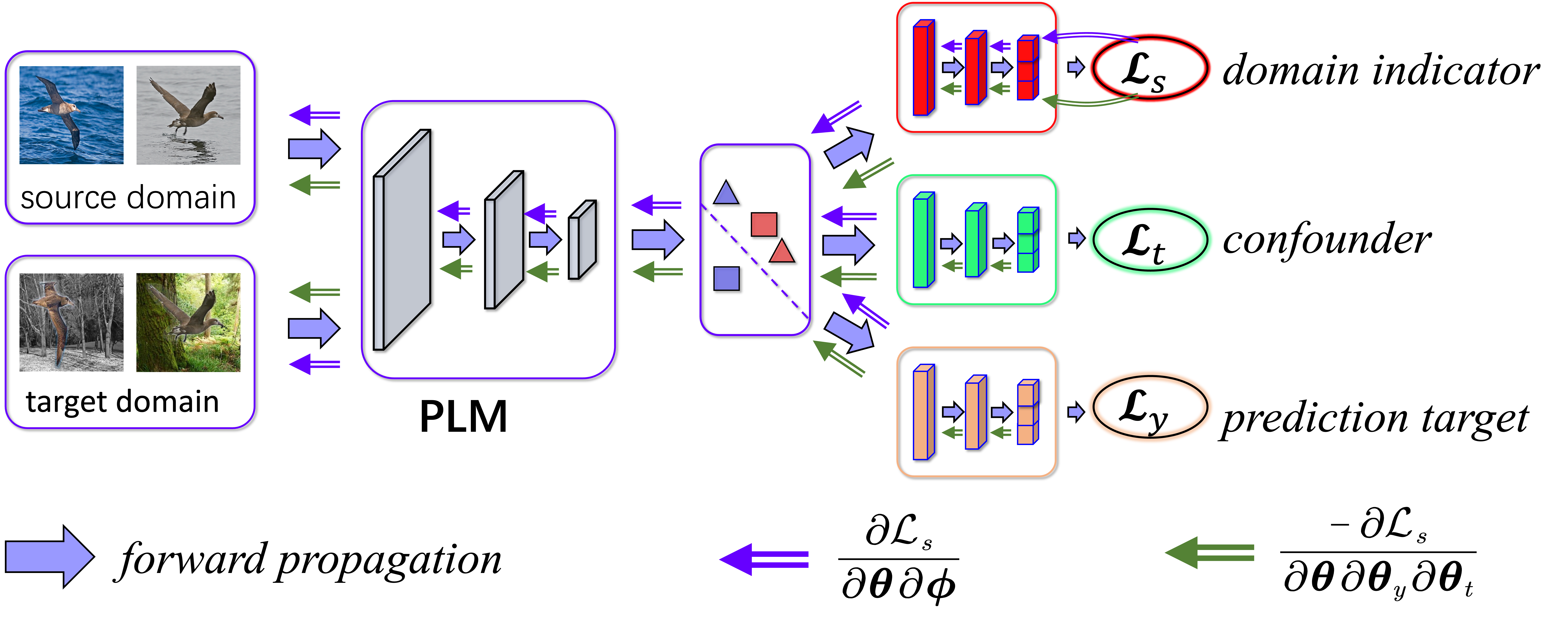}
    \caption{The schema of the CadaFT framework, where some input features $\bm{x}^{\star}$, $t$, $t^{\star}$, $y$, $y^{\star}$, $s^{\star}$, and $s$ can be masked in some tasks for single-source domain adaptation.}
    \label{afrel}
\end{figure}

The CadaFT framework is set as a latent variable model,
where each instance $\mathcal{I}^{(n)}$ has a latent variable $\bm{z}^{(n)} \in \mathbb{R}^l$,
$l > 1$,
to learn the hidden representations from observed feature $\bm{z}^{(n)}=f(\bm{x}^{(n)};\boldsymbol{\theta}^z)$,
where the foundation model $f\left( \bm{x}; \boldsymbol{\theta}^z \right)$ can be a LLM or PVM.
The confounder $t_i$,
$i=1,\cdots,T$,
domain indicator $s$ and downstream task targets $\bm{y}$ are learned from discriminators $t_i^{(n)}=f\left(\bm{z}^{(n)}; \boldsymbol{\theta}^t_i \right)$,
$s^{(n)}=f\left(\bm{z}^{(n)}; \boldsymbol{\phi} \right)$,
and $y^{(n)}=f\left(\bm{z}^{(n)}; \boldsymbol{\theta}^y \right)$ respectively.
The implementation schema of CadaFT is summarized in Figure \ref{afrel}.

The above CadaFT framework introduces a confounder controlling unit on top of the traditional adversarial domain adaptation models.
As a result,
CadaFT further balances the confounding biases in the representation space beyond the domain-level representations,
thereby enhancing the model's domain generalization ability.
In addition,
CadaFT balances the domain discrepancy in representations via adversarial learning,
which can draw the distribution contours for all domains into the same sub-space.
This way,
all source and target domain features are utilized to debias the confounders.

\subsection{Domain Classification with Confounder Balancing}

The input data $\lbrace \mathcal{I}; \mathcal{I}^{\ast} \rbrace$ includes all source domain $\mathcal{I}$ and target domain data $\mathcal{I}^{\ast}$ (in most cases, the few-shot examples),
where the prediction target $y^{\ast} \sim \mathcal{I}^{\ast}$ for target domains is unavailable.
CadaFT implicitly minimizes the averaged treatment effect on confounders over all domains.
Given the identified confounders,
the domain discrepancy has $2^{T \times S}$ comparing pairs,
where $T$ is the number of confounders and $S$ is the number of domains.

The domain discrepancy estimator is defined as follows,

\begin{equation}
    \ell \left( s, t, s^{\star}, t^{\star} \right) = \left| \mathbb{E} \left[ p_{\bm{z} \sim q(\bm{z}|\bm{x})} \left( s|\bm{z}, t \right) \right] - \mathbb{E} \left[ p_{\bm{z}^{\star} \sim q(\bm{z}^{\star}|\bm{x}^{\star})} \left( s^{\star}|\bm{z}^{\star}, t^{\star} \right) \right] \right|
\end{equation}

\begin{figure}[htbp]
    \centering
    \includegraphics[width=10cm]{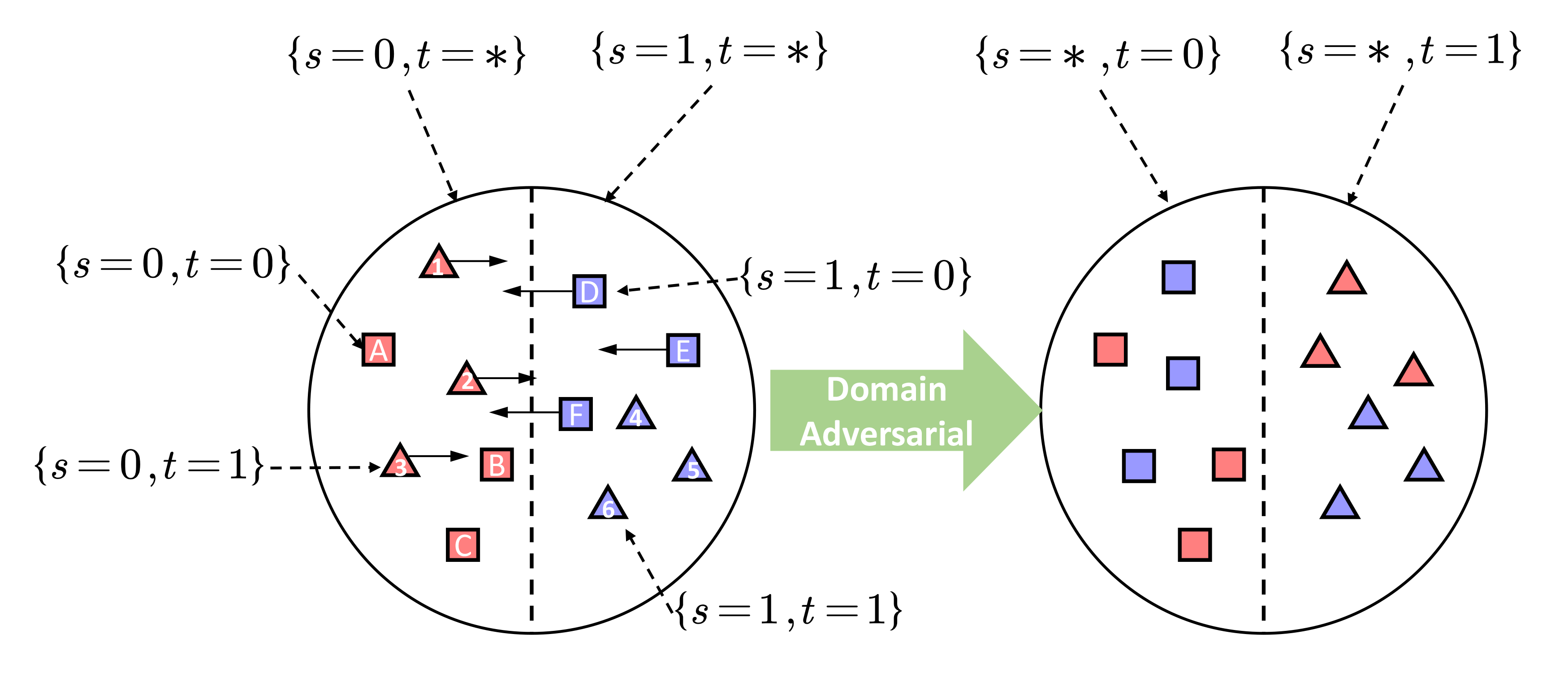}
    \caption{An example about the disentangling of domain and confounders.}
    \label{fig:domainadv}
\end{figure}

The confounder balanced adversarial domain classification can be defined as a \textbf{Minimax} problem \cite{lei2021stability}.
It aims to disentangle the spurious correlations between domains and confounders,
as shown in Figure \ref{fig:domainadv}.
Its optimization objective is defined as follows,

\begin{equation}
    \centering
    \begin{aligned}
        \underset{\boldsymbol{\phi }}{\min}\,\,\underset{\boldsymbol{\theta}^z, \boldsymbol{\theta}^t_1, \cdots, \boldsymbol{\theta}^t_T, \boldsymbol{\phi}, \boldsymbol{\theta}^y }{\max} \lbrace & \mathbb{E} _{\boldsymbol{z}\sim q\left( \boldsymbol{z}|\boldsymbol{x} \right)}\left[ \log p\left( s|\boldsymbol{z},t; \boldsymbol{\theta}^z, \boldsymbol{\theta}^t_1, \cdots, \boldsymbol{\theta}^t_T, \boldsymbol{\phi}, \boldsymbol{\theta}^y \right) \right] \\
  & +\mathbb{E} _{\boldsymbol{z}^{\ast}\sim q\left( \boldsymbol{z}|\boldsymbol{x}^{\ast} \right)}\left[ \log p\left( s^{\ast}|\boldsymbol{z}^{\ast},t^{\ast};\boldsymbol{\theta}^z, \boldsymbol{\theta}^t_1, \cdots, \boldsymbol{\theta}^t_T, \boldsymbol{\phi}, \boldsymbol{\theta}^y \right) \right] \\
    \end{aligned}
\end{equation}
where $\boldsymbol{\theta}^z, \boldsymbol{\theta}^t_1, \cdots, \boldsymbol{\theta}^t_T, \boldsymbol{\phi}, \boldsymbol{\theta}^y$ represent the trainable parameters in foundation model,
domain and confounder discriminators.

\begin{algorithm}
\caption{CadaFT framework}
\label{alg:dread}
\textbf{Input}: ID data $\mathcal{I} = \lbrace \bm{x}^{(i)}, y^{(i)}, t^{(i)}, s^{(i)} \rbrace ^{N}_{i=1}$, OOD data $\mathcal{I}^{\ast} = \lbrace \bm{x}^{\ast (i)}, y^{\ast (i)}, t^{\ast (i)}, s^{\ast (i)} \rbrace ^{N}_{i=1}$.\\
\textbf{Parameter}: Training epochs $E$, $T$ for discriminator, mini-batch size $B$ and learning rate $r$.\\
\textbf{Output}: Predicted targets $\widehat{y}^{\star}$.
\begin{algorithmic}[1] 
\State \textbf{Randomly initialize} $\boldsymbol{\theta}^z, \boldsymbol{\theta}^t_1, \cdots, \boldsymbol{\theta}^t_T, \boldsymbol{\phi}, \boldsymbol{\theta}^y$;
\While{$e \leftarrow 1$ to $E$}
\State \# adversarial training step
\For{$t \leftarrow 1$ to $T$}
\State \# sample mini-batches
\State sample ID data $\lbrace \left(\bm{x}^{(i)}, s^{(i)} \right) \rbrace ^{B}_{i=1} \sim \mathcal{I}$, $s^{(i)} \sim \bm{U} \left( N, 1/N \right)$
\State sample OOD data $\lbrace \bm{x}^{\ast (i)}, s^{\ast (i)} \rbrace ^{B}_{i=1} \sim \mathcal{I}^{\ast}$, $s^{\ast (i)} \sim \bm{U} \left( N, 1/N \right)$
\State update the parameters $\boldsymbol{\phi}$: $\frac{\partial}{\partial \boldsymbol{\phi }}\left[ \mathcal{L}_{\min} \right] $; \quad \# Eq.\ref{maxopt}
\EndFor
\State \# updating $\boldsymbol{\theta}^z, \boldsymbol{\theta}^t_1, \cdots, \boldsymbol{\theta}^t_T, \boldsymbol{\phi}, \boldsymbol{\theta}^y$
\State \# feature extractor training step
\State \# sample mini-batches
\State sample Domain-1 data $\lbrace \bm{x}^{(i)}, y^{(i)}, s^{(i)}, t^{(i)} \rbrace^{B}_{i=1} \sim \mathcal{I}$
\State sample Domain-2 batch $ \lbrace \bm{x}^{\ast (i)}, s^{\ast (i)}, t^{\ast (i)} \rbrace^{B}_{i=1} \sim \mathcal{I}^{\ast}$
\State update parameters $\boldsymbol{\theta}^z, \boldsymbol{\theta}^t_1, \cdots, \boldsymbol{\theta}^t_T, \boldsymbol{\phi}, \boldsymbol{\theta}^y$: $\frac{\partial}{\partial \boldsymbol{\phi }}\left[ \mathcal{L}_{\max} \right] $; \quad \# Eq.\ref{minopt}
\EndWhile
\State \textbf{return} $\widehat{y}$
\end{algorithmic}
\end{algorithm}

\subsection{Adversarial Loss Function}

We use the adversarial loss function to implement the objective in the above Minimax game with two steps.
(\romannumeral1) The feature extractor training is supervised by the task labels and confounders from the source domain and (\romannumeral2) the adversarial training by minimizing the ability of the domain classifier.

\textbf{Maximization} step extracts feature $\bm{z}$ from the instance feature $\bm{x}$ supervised by the label $y$, confounder $\bm{t}$ and domain $s$.
The domain classifier is semi-supervised as the target domain $s^{\star}$ in $\mathcal{I}^{\ast}$ is masked in training.
And the optimization objective is $\underset{\boldsymbol{\Theta}}{\min}\mathcal{L}_{FE}$,

\begin{equation}
    \centering
    \begin{aligned}
        & \mathcal{L}_{\min} = \mathbb{E}_{\left( \bm{x}, \bm{s}, \bm{t}, \bm{y} \right) \sim \mathcal{I}} \lbrack - \left(\bm{y} \log \widehat{\bm{y}} + \left( \bm{1}-\bm{y} \right) \log \left( \bm{1} - \widehat{\bm{y}} \right) \right) \\
        & \qquad \qquad \qquad \qquad - \left( \bm{s} \log \widehat{\bm{s}} + \left( \bm{1} - \bm{s} \right) \log \left( \bm{1} - \widehat{\bm{s}} \right) \right) \\
        & \qquad \qquad \qquad \qquad - \left( \bm{t} \log \widehat{\bm{t}} + \left( \bm{1} - \bm{t} \right) \log \left( \bm{1} - \widehat{\bm{t}} \right) \right) \rbrack \\
        & \quad \qquad +\mathbb{E} _{\left( \bm{x}^{\ast}, \bm{s}^{\ast}, \bm{y}^{\ast} \right) \sim \mathcal{I} ^{\ast}} \lbrack - \left( \bm{s}^{\ast} \log \widehat{\bm{s}}^{\ast} + \left( \bm{1} - \bm{s}^{\ast} \right) \log \left( \bm{1} - \widehat{\bm{s}^{\ast}} \right) \right) \\
        & \qquad \qquad \qquad \qquad - \left( \bm{t} \log \widehat{\bm{t}}^{\ast} \right) + \left( \bm{1} - \bm{t}^{\ast} \right) \log \left( \bm{1} - \widehat{\bm{t}^{\ast}} \right) \rbrack \rbrace \\
        & \quad \mbox{where} \quad \bm{z}\sim p_{\boldsymbol{\theta}^z}\left( \bm{z}|\bm{x} \right), \widehat{\bm{y}} \sim p_{\boldsymbol{\theta}^y} \left( \bm{y}|\bm{z} \right) ,\widehat{\bm{s}}\sim p_{\boldsymbol{\phi}}\left( \bm{s}|\bm{z} \right) \\
        & \qquad \qquad \quad \widehat{\bm{t}} \sim p_{\boldsymbol{\theta}^t}\left( \bm{t}|\bm{z} \right), \bm{z}^{\ast} \sim p_{\boldsymbol{\theta}^z}\left( \bm{z}|\bm{x}^{\ast} \right), \\
        & \qquad \qquad \quad \widehat{\bm{s}^{\ast}} \sim p_{\boldsymbol{\phi}}\left( \bm{s}|\bm{z}^{\ast} \right), \widehat{\bm{t}^{\ast}} \sim p_{\boldsymbol{\theta}^t}\left( \bm{t}|\bm{z}^{\ast} \right) 
    \end{aligned}
    \label{maxopt}
\end{equation}

\textbf{Minimization} step aims to minimize the ability of the domain classifier under the proper control of confounders,
which is an adversarial learning method with a competing objective on the domain label $s$.
This step minimizes the predicted confidence for the accurate domain indicator.
The domain discrepancy is minimized in feature space $p(\bm{z})$ when the optimization objective converges $\mathcal{L}_{Adv} \rightarrow 0$.
In addition,
the confounder in the source domain is correctly predicted in the feature extractor learning step,
which is equal to the control for confounders in causal analysis.
In this way,
the bias caused by confounders can be removed. 

\begin{equation}
    \centering
    \begin{aligned}
        \mathcal{L}_{Adv} =& \mathbb{E}_{\bm{x} \sim \mathcal{I}} \left[ - \left( \widetilde{\bm{s}} \log \widehat{\bm{s}} + \left( \bm{1} - \widetilde{\bm{s}} \right) \log \left( \bm{1} - \widehat{\bm{s}} \right) \right) \right] \\
        & + \mathbb{E} _{\bm{x}^{\ast} \sim \mathcal{I}^{\ast}} \left[ - \left( \widetilde{\bm{s}^{\ast}} \log \widehat{\bm{s}^{\ast}} + \left(\bm{1}- \widetilde{\bm{s}^{\ast}} \right) \log \left( \bm{1} - \widehat{\bm{s}}^{\ast} \right) \right) \right] \\
        & \mbox{where} \quad \widetilde{\bm{s}}, \widetilde{\bm{s}^{\ast}} \sim \bm{U} \left( N, \frac{1}{N} \right), \bm{z} \sim p_{\boldsymbol{\theta}^z}\left( \bm{z}|\bm{x} \right) ,\widehat{\bm{s}}\sim p_{\boldsymbol{\phi}}\left( \bm{s}|\bm{z} \right) \\
        & \qquad \qquad \bm{z}^{\ast}\sim p_{\boldsymbol{\theta}^z}\left( \bm{z}|\bm{x}^{\ast} \right), \widehat{\bm{s}^{\ast}} \sim p_{\boldsymbol{\phi}}\left( \bm{s}|\bm{z}^{\ast} \right) 
    \end{aligned}
 \label{minopt}
\end{equation}

The CadaFT framework is summarized in Algorithm \ref{alg:dread}.

\section{Experiments}

This section comprehensively evaluates the proposed CadaFT framework with OOD generalization and domain adaptation tasks in both NLP and CV downstream tasks.
The first experiment evaluates the framework's ability for OOD generalization and analyses its robustness to spurious correlations.
This experiment is conducted on the natural language inference (NLI), question query pair (QQP) classification in the NLP field and objective recognition in the CV field.
The second experiment is conducted on the Office-home \cite{saenko2010adapting} and MiniDomainNet \cite{peng2019moment} datasets to evaluate the effectiveness of the domain adaptation.
Overall,
experimental results demonstrate the versatility and efficacy of the CadaFT framework across different modalities.

\subsection{OOD Generalization}

\subsubsection{Text Classification}

\begin{figure}[htbp]
    \centering
    \includegraphics[width=14cm]{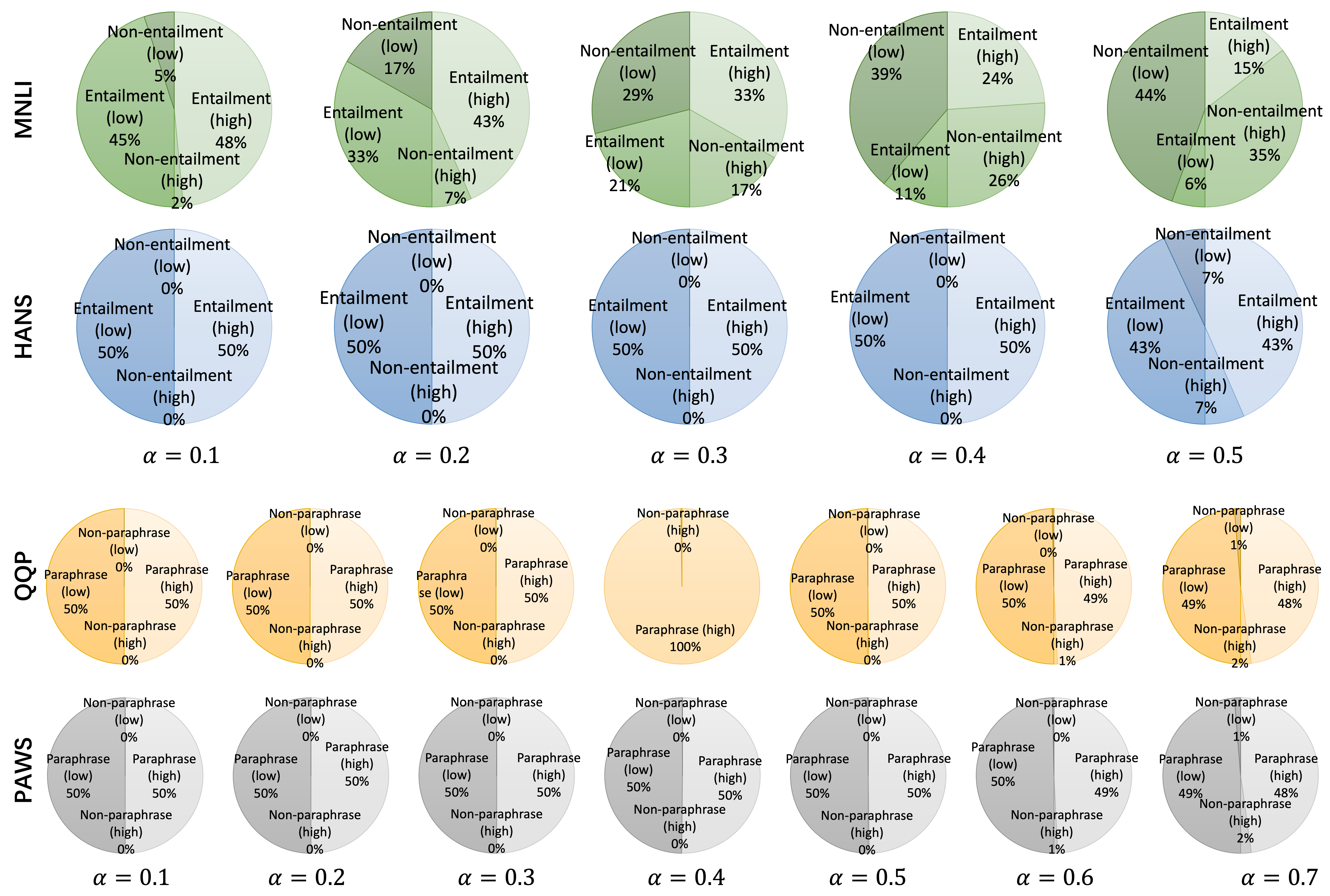}
    \caption{The statistics about the examples in different labels with high or low word overlaps under different word overlap threshold $\alpha$ in Eq.(\ref{overlapthres}). It is clear that $\alpha=0.4$, $\alpha=0.6$ are the biggest values to identify all examples as the high word overlap in HANS and PAWS respectively.}
    \label{confoundingstatistics}
\end{figure}

\textbf{Benchmarks}
For downstream NLP tasks,
the experiment is conducted on Multi-Genre Natural Language Inference (MNLI) \cite{williams2018broad} and QQP \cite{iyer2017first} datasets as the in-distribution datasets separately.
The trained models evaluate OOD generalization on HANS \cite{mccoy2020right} and PAWS \cite{zhang2019paws} respectively.
An annotated confounder in two in-distribution datasets is the word overlap.
Specifically,
as shown in Figure \ref{confoundingstatistics},
the ``entailment'' examples in MNLI have a substantially higher word overlap than other ``non-entailment'' examples \cite{rajaee2022looking}.
Instead,
HANS has high word overlap in both ``entailment'' and ``non-entailment'' examples with similar percentages \cite{mccoy2020right}. 
Similarly,
the ``paraphrase'' examples in QQP have higher word overlap than the ``non-paraphrase'' examples.
In contrast,
high word overlap is observed in both ``paraphrase'' and ``non-paraphrase'' examples in PAWS with similar percentages.

\textbf{Confounder definition.}
In both MNLI and QQP tasks,
the word overlapping percentages in paired sentences are used as a criterion to annotate confounder $t=1$ or $t=0$.
Given a sentence pair $\bm{x}_1=\lbrace x_{1,1}, x_{1,2}, \dots, x_{1,M} \rbrace$ and $\bm{x}_2=\lbrace x_{2,1}, x_{2,2}, \dots, x_{2,N} \rbrace$,
the value of the confounder is calculated as,

\begin{equation}
    t = \mathbb{I} \left( p\left( \frac{\sum^M_{i=1} \sum^N_{j=1} \mathbb{I}\left( w_{1,i} = w_{2,j} \right) }{M+N} \right) \geq \alpha \right)
    \label{overlapthres}
\end{equation}
where the threshold $\alpha$ is set as a value $0.4$ for MNLI and $0.6$ for QQP.
This threshold guarantees all examples in the OOD dataset - HANS, PAWS are labeled as high word overlap as the statistics shown in Figure \ref{confoundingstatistics}.

\textbf{Compared Baselines}
\begin{enumerate}
    \item \textbf{LLMs:} BERT \cite{kenton2019bert}, RoBERTa \cite{liu2019roberta} and LLaMA\cite{meta2023introducing}, LLaMA-2, OPT, GPT-3.5, GPT-4.
    \item The \textbf{prompt tuning (Prompt-T)} for LLMs \cite{liu2023pre, utama2021avoiding} does not fine-tune a model on a labeled dataset for a specific task. Instead, they instead use the LLMs' existing knowledge and adjust the input prompts to get the desired output.
    \item \textbf{Regularization fine-tuned (RegFT)} LLMs \cite{utama2021avoiding} employ a regularization technique to retain the pre-training weights.
    This approach has been proven effective in few-shot fine-tuning,
    as it prevents the erasure of valuable knowledge from pre-training.
    \item Zero-shot prompting GPT-3, GPT-3.5 and GPT-4 via OpenAI's official API \cite{ye2023comprehensive, wang2023robustness} .
    \item \textbf{The debiasing methods:} \textbf{BERT}-$\mathcal{F}$ \cite{yaghoobzadeh2021increasing} fine-tunes the BERT with forgotten examples at each training epoch. The forgotten examples in a model are correctly classified at some point and are misclassified in the following training.
    \textbf{ReWeighting} \cite{clark2019don} involves training a naive model to predict based on dataset biases and another, robust model in ensemble with it, encouraging the latter to focus on more generalizable data patterns.
    The confidence regularization method (\textbf{Reg-conf}) \cite{utama2020mind} is designed to provide sufficient incentives for models to learn invariant representations from all training data.
    \textbf{Z-filtering} is a data augmentation that trains data generators to produce high-quality, label-consistent samples while removing spurious correlations.
    The \textbf{product-of-experts (POE)} \cite{mahabadi2020end} combines the probability distributions of the bias-only and the foundation model to make predictions based on different input characteristics. Next, the base model is trained using the cross-entropy loss of the combined classifier.
    The \textbf{Learned-Mixin (Lmin)} \cite{clark2019don} method is a two-stage approach that involves: (1) training a naive model with prior knowledge of dataset biases and (2) training a robust model with the naive model in an ensemble to focus on patterns that are more likely to generalize.
    \textbf{AdaTest} employs LLMs and human feedback to autonomously generate unit tests, identifying bugs in a target model. These are then rectified through an iterative text-fix-retest cycle, mirroring conventional software development practices.
    \textbf{SBERT} and \textbf{SRoBERTa} utilize a biencoder approach to explore the effects of explicitly integrating predicate-argument information via weighted aggregation.
\end{enumerate}

\textbf{Implementation details} \\
The feature extractor $f\left( \bm{x}; \boldsymbol{\theta}^z \right)$ in CadaFT is implemented with BERT, RoBERTa and LLaMA respectively.
CadaFT has trained $20$ epochs on 4$\times$V100 GPU with Adam optimizer.
The initial learning rate is 2e-5,
the L2 weight decay is $0.01$, and the mini-batch size is set to $32$.

\begin{table}[htbp]
  \centering
    \begin{tabular}{lcccc}
    \toprule
    \multicolumn{1}{c}{\multirow{3}[4]{*}{Models}} & \multicolumn{2}{c}{NLI} & \multicolumn{2}{c}{PI} \\
\cmidrule(r){2-3} \cmidrule(r){4-5}
& ID & OOD & ID & OOD \\ \cmidrule(r){2-3} \cmidrule(r){4-5}
& dev-matched & HANS & QQP & PAWS \\
    \midrule
    BERT-base \cite{yaghoobzadeh2021increasing} & 67.9 & 49.9 & 83.0 & 40.6 \\
    BERT-base+$\mathcal{F}_{\text{BoW}}$ \cite{yaghoobzadeh2021increasing} & 83.1 & 70.5 & 89.0 & 48.8 \\
    BERT-base+$\mathcal{F}_{\text{BiLSTM}}$ \cite{yaghoobzadeh2021increasing} & 82.9 & 70.4 & 88.0 & 47.6 \\
    RoBERTa-large & 89.1 & 77.1 & 89.0 & 39.5 \\
    ReWeighting \cite{clark2019don} & 83.5 & 69.2 & 83.5  & 69.2 \\
    Reg-conf \cite{utama2020mind} & 84.3 & 69.1 & 89.1  & 39.8 \\
    PoE \cite{clark2019don}  & 84.0 & 66.5 & 89.2  & 55.2 \\
    Lmin \cite{clark2019don} & 84.3 & 64.0 & - & - \\
    AdaTest \cite{ribeiro2022adaptive} & - & - & 91.1 & 53.8 \\
    SBERT \cite{peng-etal-2022-predicate} & - & - & 90.8 & 66.0 \\
    SRoBERTa \cite{peng-etal-2022-predicate} & - & - & 90.8 & 67.4 \\
    Debiasing Masks \cite{meissner2022debiasing} & 81.9 & 68.7 & 89.6 & 44.3 \\
    RegFT (few-shot) \cite{utama2021avoiding} & 82.7  & 60.2  & 81.5  & 37.1 \\
    Prompt-T$^\clubsuit$ (few-shot \#512) \cite{liu2023pre} & 84.3  & 54.8  & 82.1  & 29.6 \\
    Prompt-T$^\clubsuit$ (zero-shot) \cite{liu2023pre} & 51.1  & 62.6  & 35.4  & 51.8 \\
    Prompting GPT-3$^{\star}$ \cite{si2022prompting} & 56.8 & 75.3 & 91.3 & 40.1 \\
    Prompting GPT-3.5$^\spadesuit$ (few-shot \# 16) & 54.6 & 66.8 & 72.4 & 42.4 \\
    OPT-13B & 85.5 & 70.8 & 91.2 & 47.5 \\
    LLaMA-13B & 85.3 & 75.3 & 90.5 & 46.9 \\
    LLaMA2-13B & 87.3 & 70.7 & 90.7 & 69.2 \\
    Prompting GPT-4$^\spadesuit$ (zero-shot) & 38.4 & 60.8 & 76.8 & 69.4 \\
    Prompting GPT-4$^\spadesuit$ (few-shot \#16) & 49.5 & 60.0 & 91.6 & 68.7 \\
    \midrule
    CadaFT$^\clubsuit$ & 88.5 & 80.2 & 91.6 & 51.5 \\
    CadaFT$^\clubsuit$ (few-shot \#16) & \textbf{90.2} & \textbf{85.6} & \textbf{95.6} & \textbf{81.5} \\
    \bottomrule
    \end{tabular}%
    \caption{Averaged test accuracy ($\%$) over 5 runs on MNLI and QQP and corresponding OOD evaluations, where $^\clubsuit$ labeled methods take the RoBERTa-large as their foundation models, $\star$ is tested with the prompt on the GPT-3 version of text-Davinci-002 and$^\spadesuit$ are tested with the prompts on GPT-3.5-turbo and GPT-4.}
  \label{tab:mainresults}%
\end{table}%

\textbf{Main Results}\\
\begin{figure}[htbp]
    \centering
    \includegraphics[width=14cm]{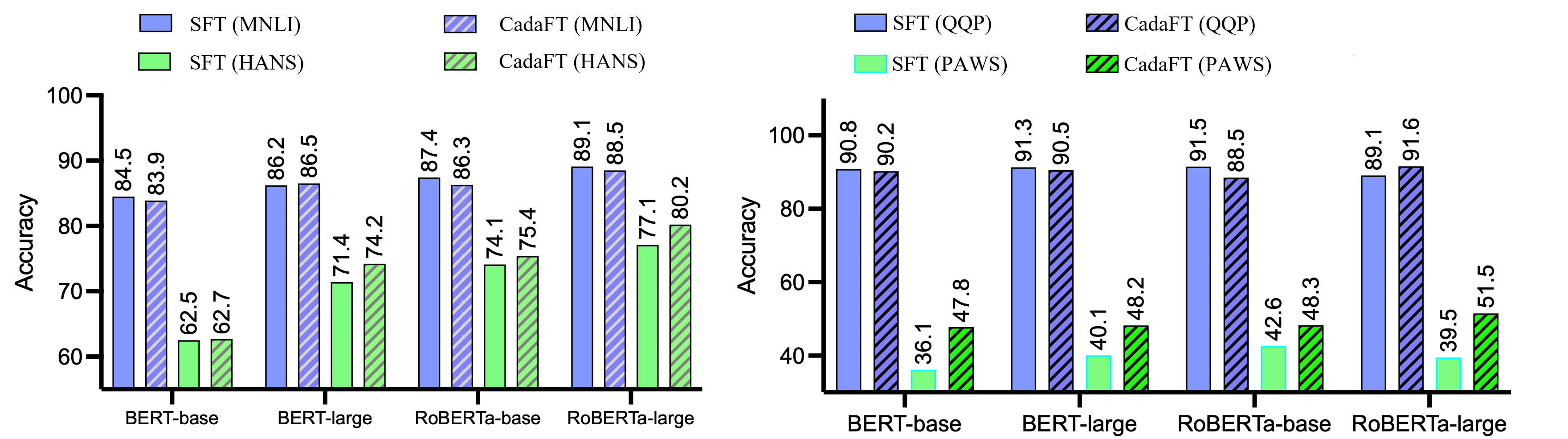}
    \caption{Test results of CadaFT fine-tuned (FT) different LLMs.}
    \label{fig:SPCtask}
\end{figure}

Table \ref{tab:mainresults} displays the test accuracy of compared baselines on in-distribution datasets - MNLI, QQP and corresponding OOD datasets - HANS and PAWS.
Among all compared baselines,
CadaFT (fine-tuning with few-shot OOD examples) beats other baselines on four evaluations.
Specifically,
CadaFT (few-shot) raises the test accuracies by at least $4.9\%$, $10.3\%$, $5.1\%$ and $34.6\%$ than their foundation model (LLaMA-13B) respectively,
and also raises the previous SOTA results by $1.1\%$, $8.5\%$, $4.0\%$ and $12.2\%$ respectively.
Even though without the few-shot OOD examples,
CadaFT also achieves the leading level of performance on MNLI, HANS and QQP datasets.
Notably,
CadaFT beats the newest LLMs - GPT-3, GPT-3.5, GPT-4, OPT, LLaMA and LLaMA-2 on all evaluations.
Additionally,
with the few-shot OOD datasets,
GPT-3.5, GPT-4 do not improve OOD generalization,
while CadaFT shows significant raises that $1.7\%$, $5.4\%$, $4.0\%$ and $30.0\%$ on both ID and OOD evaluations.

In addition,
as shown in Figure \ref{fig:SPCtask} where different LLMs as the foundation models for the feature extractor network $f(\bm{x}; \boldsymbol{\theta}^z)$ are supervised fine-tuned (SFT) and CadaFT fine-tuned (CadaFT-FT),
CadaFT-FT LLMs outperform their counterparts on all OOD evaluations.
Specifically,
CadaFTs-FT BERT-base, BERT-large, RoBERTa-base and RoBERTa-large beat their counterparts by $0.2\%$, $2.8\%$, $1.3\%$ and $3.1\%$ on HANS respectively,
and by $11.6\%$, $8.1\%$, $5.9\%$ and $12.0\%$ on PAWS respectively.
It shows that the CadaFT framework effectively improves the OOD generalization of most pre-trained LLMs.

\textbf{Effectiveness in Few-shot Learning}

\begin{figure}[htbp]
    \centering
    \includegraphics[width=14cm]{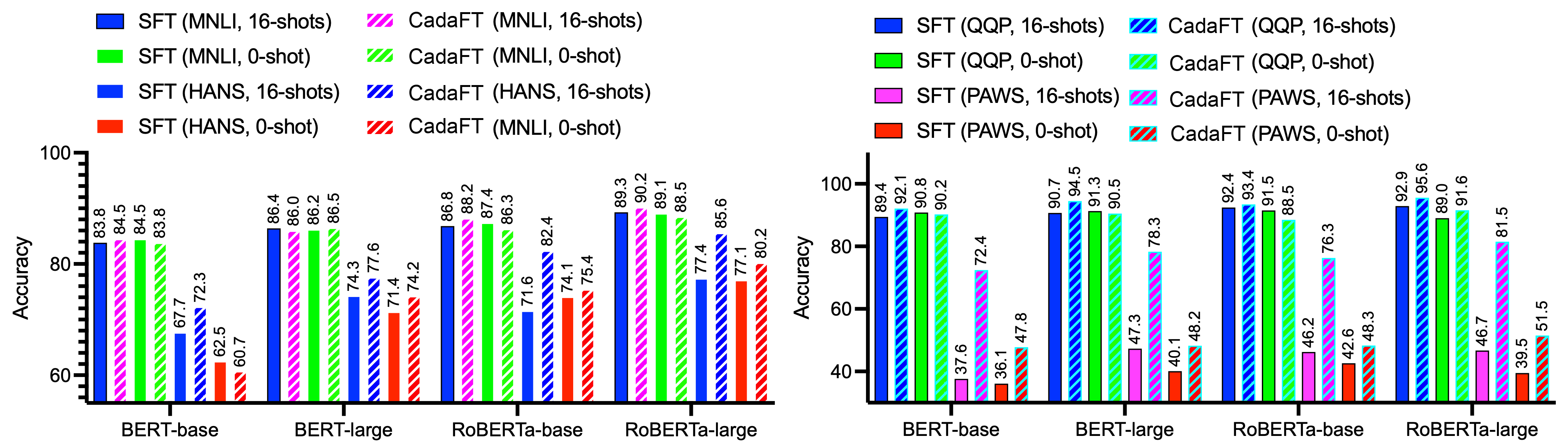}
    \caption{Averaged test accuracy over 5 runs in 16-shots OOD examples setting.}
    \label{fig:fewshoteffect}
\end{figure}

Figure \ref{fig:fewshoteffect} lists the results when LLMs are fine-tuned with few-shot OOD examples.
These results show that CadaFT fine-tuned RoBERTa-large with 16-shot OOD examples achieves the best results on all evaluations with average test accuracies of $90.2\%$ and $81.5\%$,
respectively.
Notably,
CadaFT fine-tuned RoBERTa-large with few-shot OOD examples demonstrates more significant improvements with $11.6\%$ at most on HANS and $30.1\%$ at most on PAWS.
These improvements far exceed the corresponding supervised fine-tuning counterparts - SFT RoBERTa-large with raises of $5.2\%$ at most on HANS and $7.2\%$ on PAWS.

\begin{table}[htbp]
\centering
\begin{tabular}{lcccc}
\hline
\multirow{3}{*}{Models} & \multicolumn{2}{c}{Trained on MNLI} & \multicolumn{2}{c}{Trained on QQP} \\ \cmidrule(r){2-3} \cmidrule(r){4-5}
 & ID & OOD & ID & OOD \\
 & dev-matched & HANS & QQP & PAWS \\ \hline
\multicolumn{5}{l}{\textbf{\textit{Supervised Fine-tuning LLMs}}} \\
\quad BERT-base & 84.5±0.1 & 62.5±3.4 & 90.8±0.3 & 36.1±0.8 \\
\quad BERT-large & 86.2±0.2 & 71.4±0.6 & 91.3±0.3 & 40.1±1.8 \\
\quad RoBERTa-base & 87.4±0.2 & 74.1±0.9 & 91.5±0.2 & 42.6±1.9 \\
\quad RoBERTa-large & 89.1±0.1 & 77.1±1.6 & 89.0±3.1 & 39.5±4.8 \\ \hline
\multicolumn{5}{l}{\textbf{\textit{CadaFT w/ t}}} \\
\quad BERT-base & 83.9±0.6 & 60.7±0.6 & 90.2±0.1 & 47.8±0.3 \\
\quad BERT-large & 86.5±0.1 & 74.2±0.2 & 90.5±0.3 & 48.2±0.3 \\
\quad RoBERTa-base & 86.3±0.7 & 75.4±0.3 & 88.5±0.8 & 48.3±0.2 \\
\quad RoBERTa-large & 88.5±0.2 & \textbf{80.2±0.7} & 91.6±0.3 & \textbf{51.5±0.6} \\ \hline
\multicolumn{5}{l}{CadaFT w/o t} \\
\quad BERT-base & 83.7±0.4 & 61.2±0.6 & 90.9±0.3 & 41.5±2.3 \\
\quad BERT-large & 84.9±0.3 & 58.9±1.2 & 90.6±1.4 & 44.0±0.6 \\
\quad RoBERTa-base & 87.4±0.2 & 76.3±0.4 & \textbf{91.7±1.3} & 48.5±0.6 \\
\quad RoBERTa-large & \textbf{90.2±0.3} & 79.1±0.2 & 92.4±0.4 & 49.0±0.3 \\ \hline
\end{tabular}
\caption{The ablation study to evaluate the effectiveness of confounder controoling in CadaFT.}
\label{tab:ablation}
\end{table}

Additionally,
this study conducted an ablation study to test the importance of confounder controlling in CadaFT.
The test results in Table \ref{tab:ablation} demonstrate that CadaFT without the confounder (w/o $t$) outperforms the supervised fine-tuning LLMs on both in-distribution and OOD evaluations.
Implemented with the confounder controlling (w/ $t$),
CadaFT can further improve OOD generalization.

\begin{figure}[htbp]
    \centering
    \includegraphics[width=14cm]{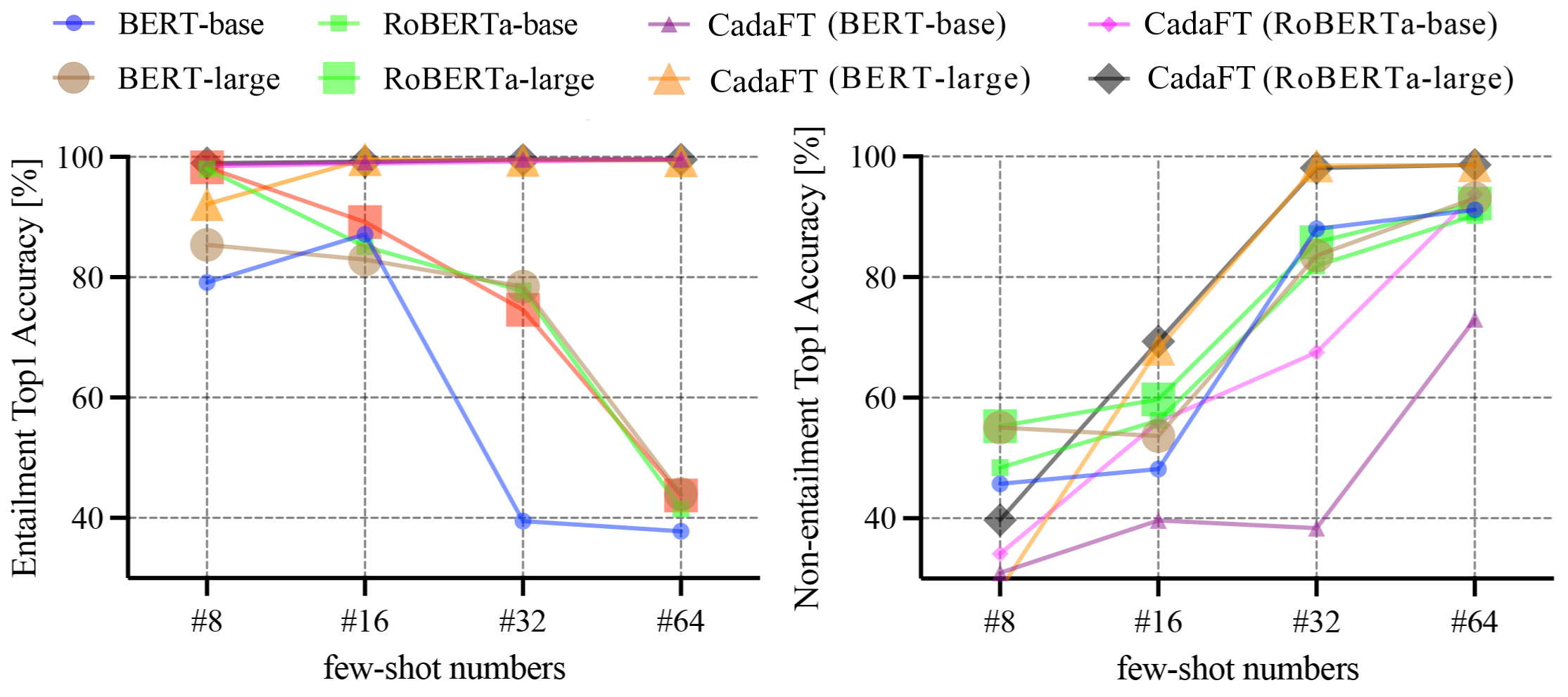}
    \caption{The top-1 OOD test accuracy ($\%$) on HANS with 5 runs with different numbers of OOD few-shot examples.}
    \label{fig:fewshotinmnli}
\end{figure}

The top-1 OOD test accuracy on HANS with different settings of few-shot are recorded in Figure \ref{fig:fewshotinmnli}.
These experimental results show that the test accuracy of non-entailment increases in fine-tuned LLMs and CadaFT fine-tuned LLMs with the rise of few-shot OOD examples.
However,
the increase of OOD few-shot examples in supervised fine-tuning LLMs dramatically impeded the test accuracy on entailment.
By contrast,
CadaFT fine-tuning can keep the performance on the entailment with the increase of OOD few-shot examples.

\begin{figure}[htbp]
    \centering
    \includegraphics[width=12cm]{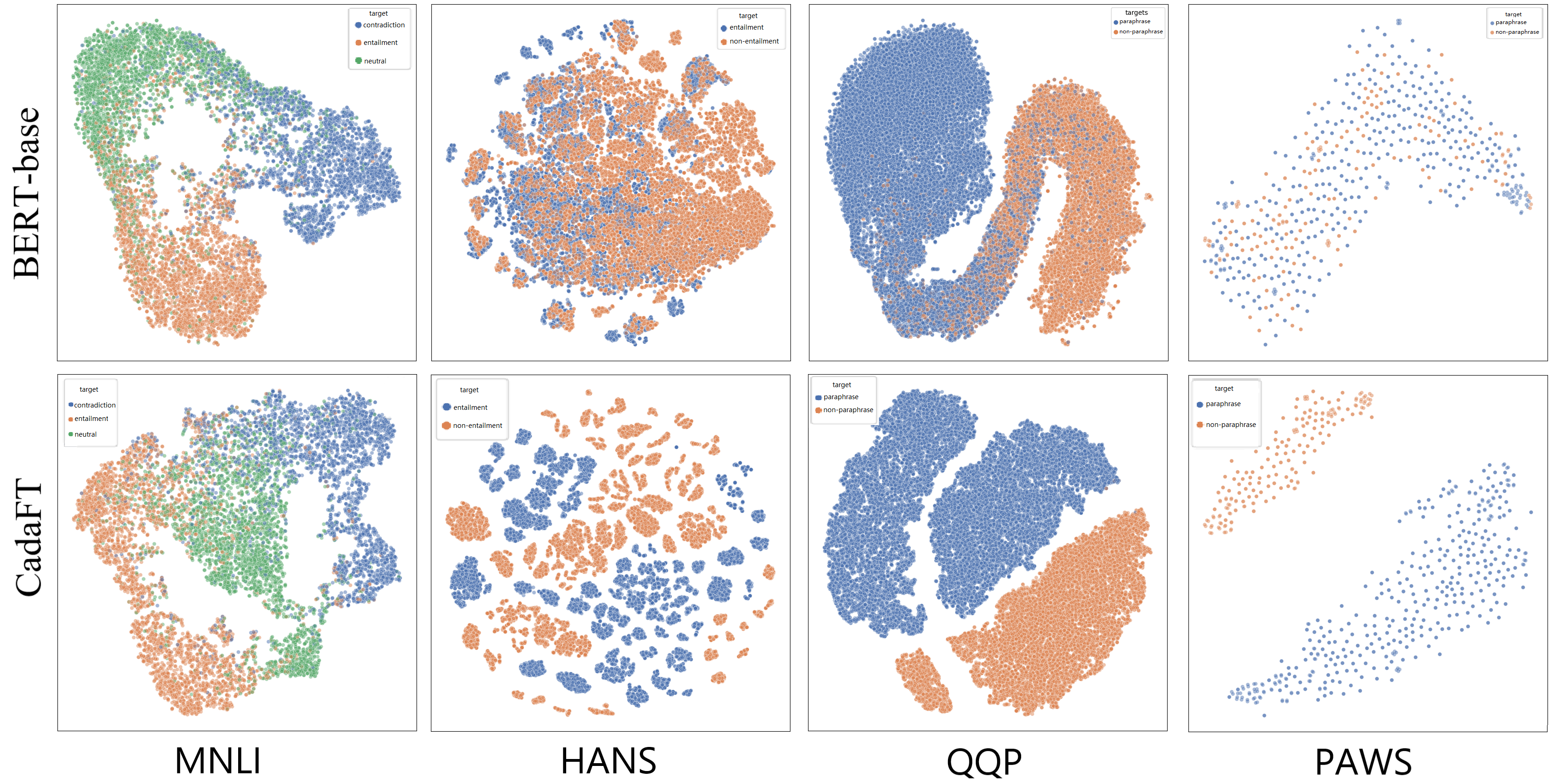}
    \caption{The t-SNE reduced representation spaces in CadaFT and fine-tuned BERT-base on MNLI, HANS, dev-QQP and PAWS.}
    \label{fig:SPC_Z}
\end{figure}

~\\
\textbf{Representation visualization:} \\
Figure \ref{fig:SPC_Z} presents the t-SNE reduced 2D representations of fine-tuned BERT-base and CadaFT on MNLI, HANS, QQP and PAWS.
CadaFT and fine-tuned BERT-base separated different labels on the ID dataset - MNLI, QQP.
However,
BERT-base learned representations on OOD data overlapped more than those learned by CadaFT.

\subsubsection{Object recognition}

~\\
\textbf{Evaluated Benchmarks:}
\begin{enumerate}
    \item \textbf{OOD evaluation with incorrect backgrounds (Waterbirds)} \\
    In object recognition tasks,
    models often rely on the background to infer object labels rather than the features of the object itself \cite{sagawa2019distributionally}.
    This study followed Sagawa et al. \cite{sagawa2019distributionally} to construct the OOD challenging dataset \textbf{Waterbirds}.
    In this OOD scenario,
    the dataset contains a predicted object label $y$ of either ``waterbird'' or ``landbird'',
    and a confounder $t$ of either ``water'' or ``land'' backgrounds.
    In the ID dataset,
    waterbirds more frequently appear in the water background,
    while landbirds appear more frequently in the land background,
    and the opposite is true for the OOD dataset.
    We used the source code from Sagawa et al. \cite{sagawa2019distributionally} to generate $11,648$ training examples,
    $11,648$ OOD examples and $11,648$ test examples.
    \item \textbf{OOD evaluation with incorrect demographics (CelebA)} \\
    The spurious correlation also demonstrates the associations between labels and demographic information like gender and ethnicity \cite{buolamwini2018gender}.
    This study followed Buolamwini et al. \cite{buolamwini2018gender} to construct \textbf{ID} and \textbf{OOD} datasets from the celebrity face dataset \textbf{CelebA} \cite{liu2015deep}.
    In this dataset,
    the prediction target is hair color $y=\lbrace \textit{blond}, \textit{dark} \rbrace$,
    and the confounder is gender $t=\lbrace \textit{male}, \textit{female} \rbrace$.
    This study split the CelebA dataset into ID and OOD domains in the experiment. 
    The \textbf{ID} domain contains only instances of $\lbrace y=\textit{blond}, t=\textit{female} \rbrace$
    and $\lbrace y=\textit{dark}, t=\textit{male} \rbrace$.
    The \textbf{OOD} domain contains a large number of instances of $\lbrace y=\textit{dark}, t=\textit{female} \rbrace$ and $\lbrace y=\textit{blond}, \textit{male} \rbrace$.
\end{enumerate}

\begin{table}[htbp]
  \centering
    \begin{tabular}{lcccccc}
    \toprule
    \multirow{2}[4]{*}{Model} & \multicolumn{3}{c}{Waterbirds} & \multicolumn{3}{c}{CelebA} \\ \cmidrule(r){2-4} \cmidrule(r){5-7}
    & ID    & OOD   & Avg.  & ID    & OOD   & Avg. \\
    \midrule
    ERM \cite{arjovsky2019invariant}  & 97.3±0.1 & 60.6±3.3 & 79.0  & 95.7±0.1 & 39.7±3.0 & 67.7 \\
    EIIL \cite{creager2021environment}  & 93.1±0.6 & 87.3±4.5 & 90.2  & 89.5±0.4 & 81.3±1.4 & 85.4 \\
    GEORGE \cite{sohoni2020no} & 95.7±0.5 & 76.2±2.1 & 86.0 & 94.6±0.2 & 53.7±1.3 & 74.2 \\
    JTT \cite{liu2021just}  & 91.7±0.8 & 88.0±0.7 & 89.9  & 87.2±1.2 & 77.8±2.0 & 82.5 \\
    BARACK \cite{sohoni2021barack} & 94.3±1.3 & 89.6±0.9 & 92.0 & 92.8±0.1 & 89.3±0.9 & 91.1 \\
    SSA \cite{namspread}  & 92.2±0.9 & 89.0±0.6 & 90.6  & 92.8±0.1 & 89.8±1.3 & 91.3 \\
    ViT-small \cite{dosovitskiy2020image} & 97.5±0.5 & 80.2±3.3 & 88.9  & 98.5±0.2 & 35.5±0.5 & 67.0 \\
    ViT-small (16-shots) & 98.0±0.6 & 88.2±0.4 & 93.1  & 85.4±0.4 & 94.5±0.2 & 78.0 \\
    Swin-small \cite{liu2021swin} & \textbf{99.1}±0.2 & 89.3±0.9 & 94.2  & \textbf{98.8}±0.3 & 33.4±0.4 & 66.1 \\
    Swim-small (16-shots) & 98.6±0.3 & 86.0±0.3 & 92.3  & 98.5±0.2 & 71.4±3.2 & 85.0 \\
    \midrule
    CadaFT-ViT & 98.2±0.2 & 88.4±1.2 & 93.3  & 94.0±0.3 & \textbf{96.4}±0.8 & 95.2 \\
    \quad - 16-shots & 97.8±0.8 & 93.2±0.7 & 95.5  & 95.0±2.5 & 95.9±2.6 & \textbf{95.5} \\
    \quad - w/o $\bm{t}$ & 97.8±0.4 & 94.4±0.8 & 96.1  & 95.5±0.6 & 92.7±2.2 & 94.1 \\
    CadaFT-Swin & 98.6±0.5 & \textbf{95.4}±0.2 & \textbf{97.0} & 95.2±0.6 & 94.4±1.2 & 94.8 \\
    \quad - 16-shots & 98.7±0.3 & 94.9±0.4 & 96.8  & 95.3±0.3 & 94.8±1.5 & 95.1 \\
    \quad - w/o $\bm{t}$ & 97.6±0.5 & 94.4±0.3 & 96.0  & 93.5±0.9 & 95.8±0.4 & 94.7 \\
    \bottomrule
    \end{tabular}%
    \caption{Test accuracies ($\%$) with standard deviations over 5 runs on Waterbirds and CelebA, where CadaFT takes ViT-small and Swim-small as the feature extractor respectively. The 16-shots means few-shot OOD examples in training, w/o $\bm{t}$ represents the ablation study without confounder.}
  \label{tab:or_main_results}%
\end{table}%

~\\
\textbf{Compared Baselines} \\
(\romannumeral1) Traditional deep learning models trained from scratch: Empirical Risk Minimization (ERM) \cite{arjovsky2019invariant}, Just Train Twice (JTT) \cite{liu2021just}, Spread Spurious Attribute (SSA) \cite{namspread}, Environment Inference for Invariant Learning (EIIL) \cite{creager2021environment}, GEORGE \cite{sohoni2020no}, BARACK \cite{sohoni2021barack} and (\romannumeral2) the pre-trained vision models (PVMs): vision Transformer (ViT) \cite{dosovitskiy2020image} and Swin Transformer (Swin) \cite{liu2021swin}.

~\\
\textbf{Implementation details}\\
The foundation model $f \left(\bm{x}; \boldsymbol{\theta}^z \right)$ in CadaFT is implemented with ViT-small and Swin-small respectively.
All compared models are trained on 4$\times$ V100 GPU by Adam optimizer \cite{jlb2015adam} with $\beta_1=0.9$,
$\beta_2=0.999$,
a low weight decay $0.01$ and a batch size $64$.
The initial learning rate $r=0.0001$ with the linear warmup and decay is used.

~\\
\textbf{Result Analysis}\\
Table \ref{tab:or_main_results} lists the test accuracy for ID and OOD evaluations of Waterbirds and CelebA tasks,
as well as the averaged accuracy (Avg.).
Even though the fine-tuned Swin-small achieved the best ID test accuracies on both Waterbirds and CelebA datasets,
it performed badly on two OOD evaluations.
On the CelebA dataset,
CadaFT fine-tuned ViT-small achieved the best OOD results among all compared baselines,
and few-shot OOD examples benefit it to attain the top-1 averaged results.
On Waterbirds,
CadaFT fine-tuned Swin-small outperformed other baselines on both OOD and averaged results.

\begin{figure}[htbp]
    \centering
    \includegraphics[width=14cm]{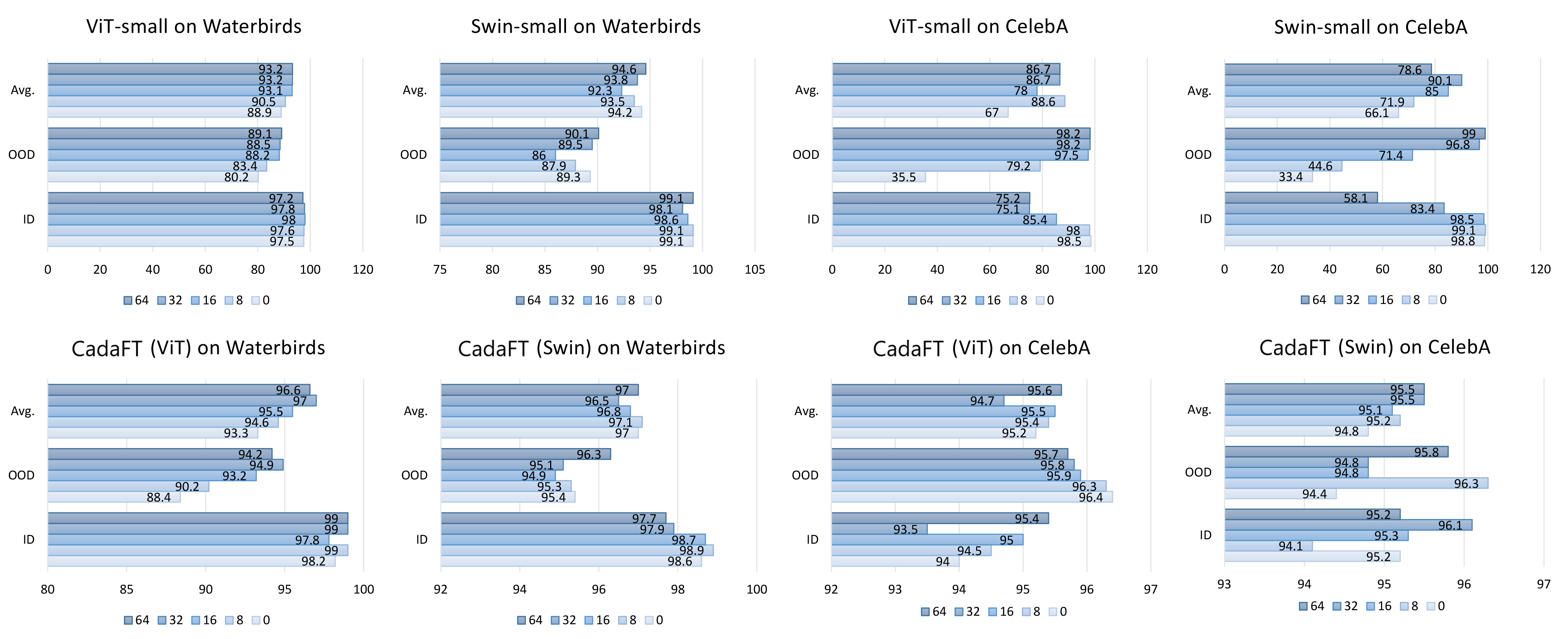}
    \caption{Test accuracy ($\%$) with different few-shot settings on Waterbirds and CelebA, where the feature extractor networks in CadaFT are implemented with ViT-small and Swim-small respectively.}
    \label{fig:or_fewshots}
\end{figure}

\textbf{Zero-shot learning}.
The proposed CadaFT significantly improved the OOD generalization,
with at least a $5.8\%$ increase on Waterbirds and a $1.9\%$ increase on CelebA.
Even though CadaFT did not show substantial improvement in ID evaluation,
it improved the averaged results over the ID and OOD test sets.

\textbf{Few-shots setting}.
Figure \ref{fig:or_fewshots} shows that fine-tuned ViT-small and Swin-small improved the OOD test accuracy by increasing the few-shot number of OOD examples,
but this OOD performance improvement is at the cost of ID test accuracy.
In contrast,
CadaFT dramatically improves the OOD generalization by increasing few-shot OOD examples while maintaining the ID performance.

\textbf{Models training from scratch},
including ERM, EIIL, GEORGE, JTT, BARACK, SSA,
demonstrate good OOD generalization even though they are inferior to the ID performance of fine-tuned PVMs,
such as the ViT-small and Swin-small.
This comparison suggests that fine-tuning PVMs on downstream tasks can destroy the pre-trained knowledge and quickly over-fit the spurious correlations on training data.

\begin{figure}[htbp]
    \centering
    \includegraphics[width=14cm]{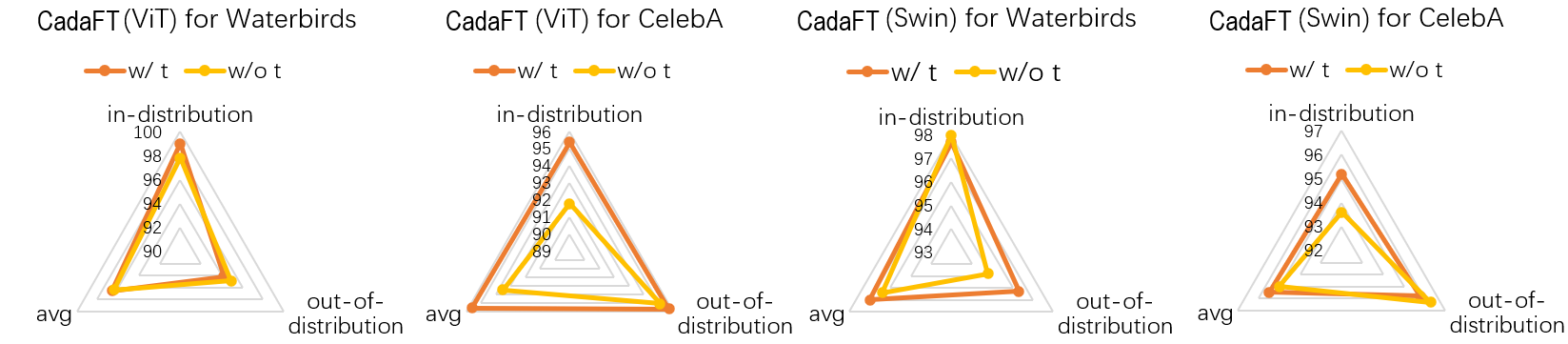}
    \caption{The ablation study demonstrates the effectiveness of the confounder classifier in CadaFT.}
    \label{fig:or_ablation}
\end{figure}

\textbf{Ablation study} Figure \ref{fig:or_ablation} conducted an ablations study to analyze the impact of the confounder controlling in CadaFT (w/o $t$).
CadaFT w/o $t$ masked the confounder $t$ in training in this experiment.
Among all compared baselines,
all CadaFT variants with $t$ outperformed the variants w/o $t$.
The radar charts in Figure \ref{fig:or_ablation} show that controlling confounders in CadaFT helps the stable performance in domain adaptation tasks.

In conclusion,
the above analysis demonstrates that the CadaFT framework is a better approach for fine-tuning PVMs and achieves better OOD generalization.
Moreover,
the confounder condition in CadaFT plays an essential role in improving OOD generalization.

\begin{figure}[htbp]
    \centering
    \includegraphics[width=12cm]{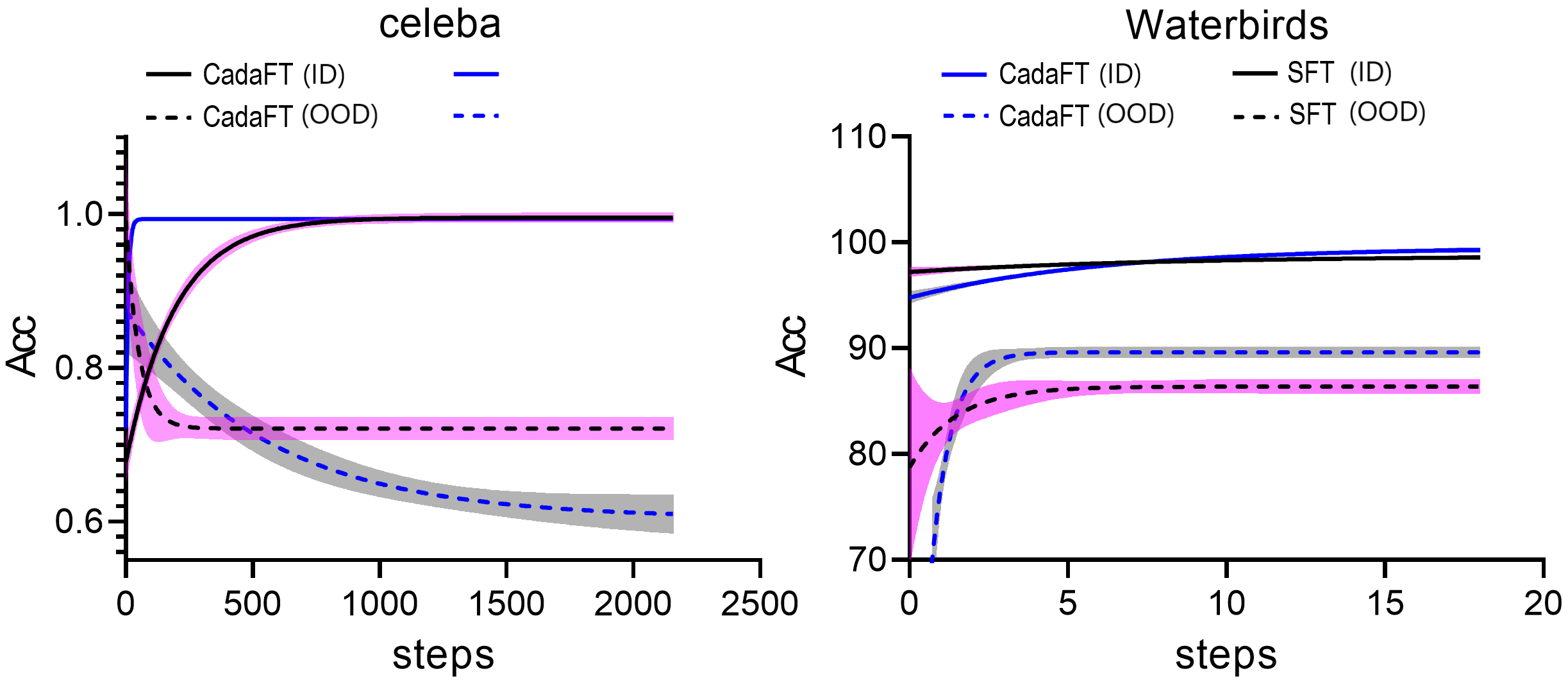}
    \caption{The test accuracies ($\%$) in training processes of ViT and CadaFT on Waterbirds and CelebA respectively.}
    \label{fig:process}
\end{figure}

\textbf{Mitigating Catastrophic Forgetting Problem}.
This study investigated the catastrophic forgetting problem in PVMs.
As shown in Figure \ref{fig:process},
although CadaFT required more training steps to converge than supervised fine-tuning ViT,
it achieves a significantly better OOD performance with $72.3\%$ than $35.5\%$ in supervised fine-tuned ViT.
On Waterbirds,
both CadaFT and supervised fine-tuned ViT keep increasing OOD generalization as the training progresses.
Nevertheless,
CadaFT takes longer to improve the OOD generalization than supervised fine-tuning ViT.
These results demonstrate that CadaFT effectively mitigates the catastrophic forgetting problem in PVMs to achieve better OOD generalization.

\begin{figure}[htbp]
    \centering
    \includegraphics[width=14cm]{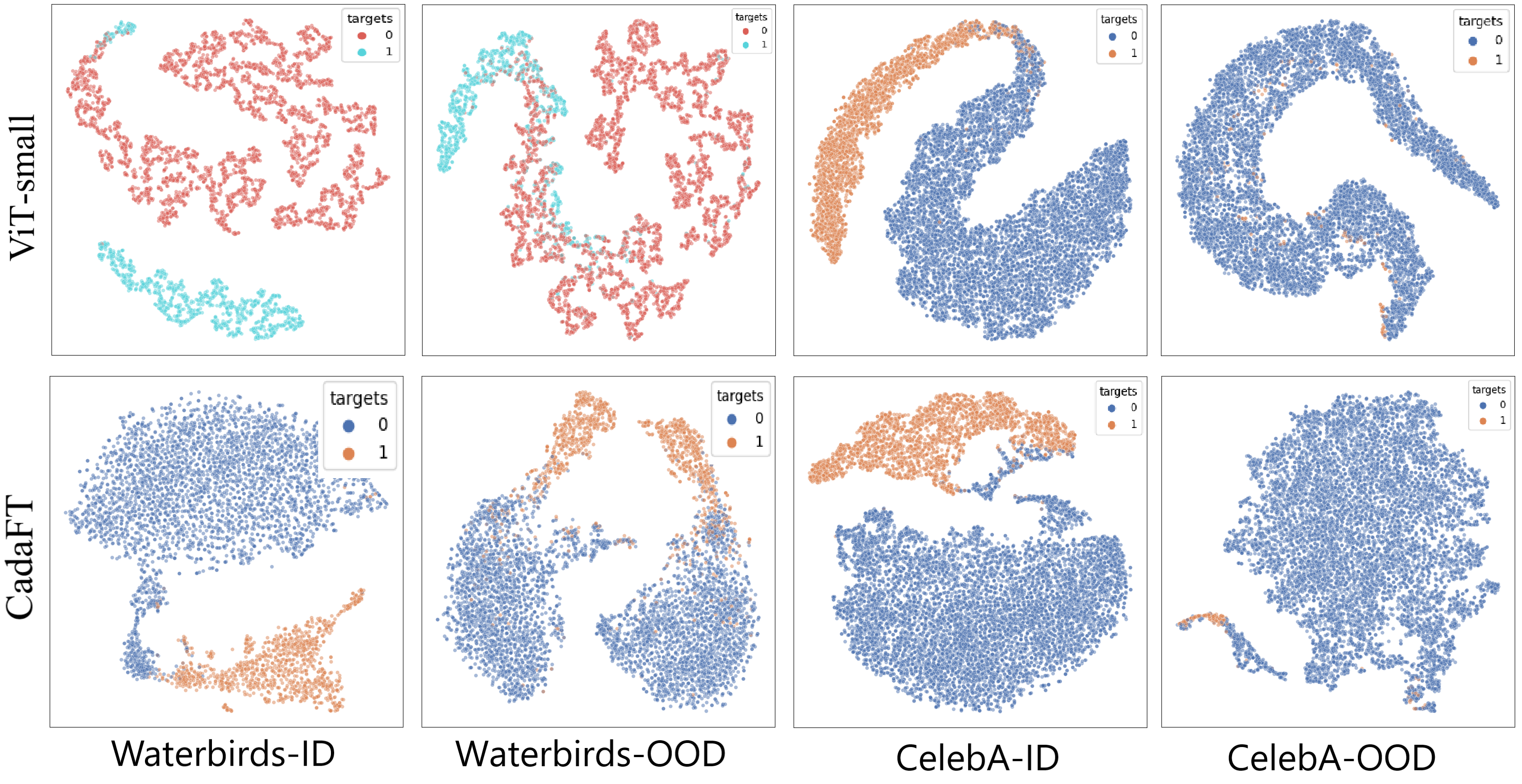}
    \caption{The t-SNE reduced representation in fine-tuned ViT-small and CadaFT (ViT-small) on ID, OOD data of Waterbirds and CelebA.}
    \label{fig:res}
\end{figure}

\textbf{Representation visualization:}
Figure \ref{fig:res} shows representation spaces reduced by t-SNE \cite{van2008visualizing} on Waterbirds and CelebA.
These results indicate that,
on ID data,
both CadaFT and fine-tuned ViT-small generate sub-spaces for different labels with clear contours.
However,
on OOD data,
the representation learned by ViT-small overlaps more sub-spaces than CadaFT.

\subsection{Domain Adaptation}

This section validates CadaFT on the single-source domain adaptation (SSDA) task of Office-home \cite{venkateswara2017deep}  and a multi-source domain adaptation (MSDA) task of MiniDomainNet \cite{zhou2021domain},
and both Office-home and MiniDomainNet are object classification tasks.
SSDA aims to adapt a single-source domain to other unseen domains,
while MSDA aims to adapt multiple source domains to an unseen domain.

\begin{figure}[htbp]
    \centering
    \includegraphics[width=12cm]{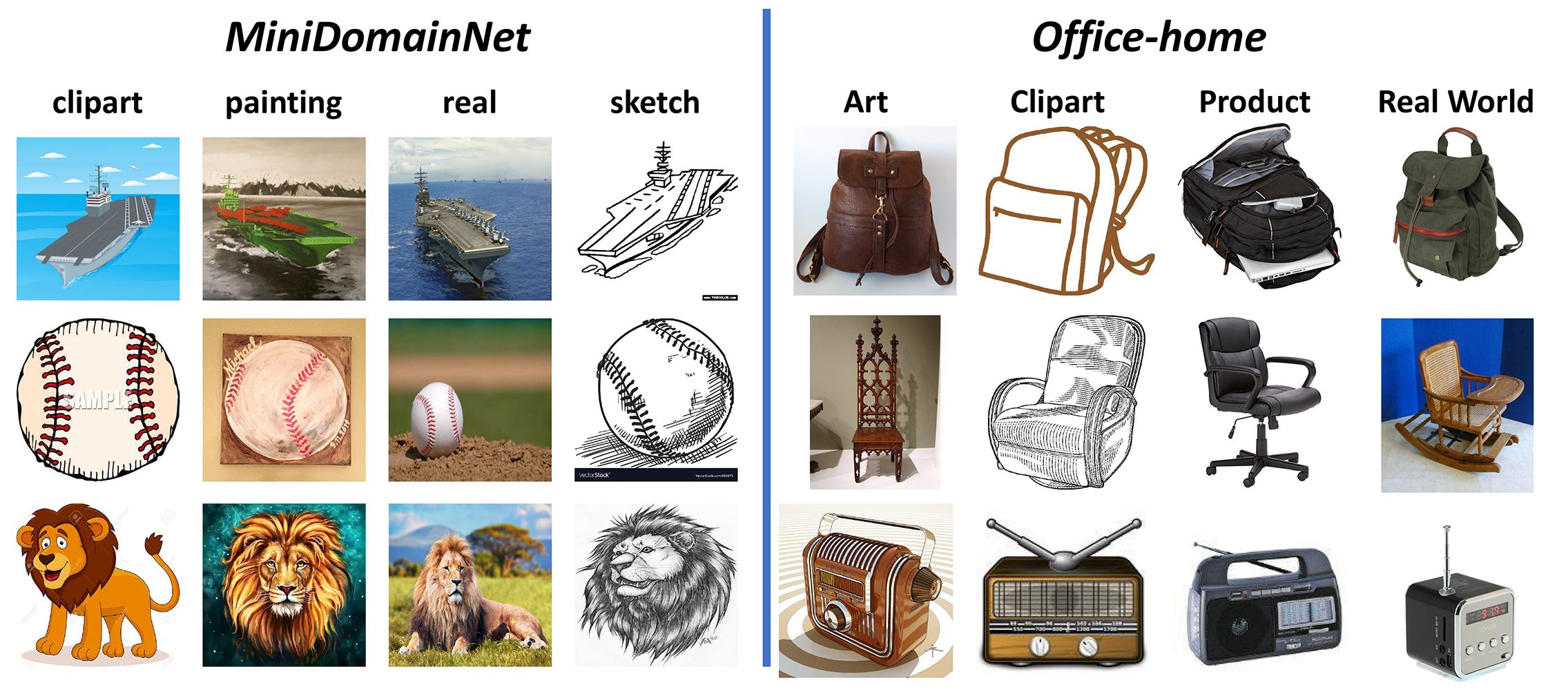}
    \caption{Illustrations of samples from four domains in Office-home and MiniDomainNet respectively.}
    \label{fig:sample_demos}
\end{figure}

\textbf{Datasets.}
Office-home is a medium-sized domain adaptation dataset with $15,500$ images collected from four domains Art ($\mathcal{A}$), Clipart ($\mathcal{C}$), Product ($\mathcal{P}$), and Real-World ($\mathcal{R}$),
and each domain contains 65 categories.
We evaluate CadaFT on 12 domain adaptation tasks: $\mathcal{A} \rightarrow \mathcal{C}$, $\mathcal{A} \rightarrow \mathcal{P}$, $\mathcal{A} \rightarrow \mathcal{R}$, $\mathcal{C} \rightarrow \mathcal{A}$, $\mathcal{C} \rightarrow \mathcal{P}$, $\mathcal{C} \rightarrow \mathcal{R}$, $\mathcal{P} \rightarrow \mathcal{A}$, $\mathcal{P} \rightarrow \mathcal{C}$, $\mathcal{P} \rightarrow \mathcal{R}$, $\mathcal{R} \rightarrow \mathcal{A}$, $\mathcal{R} \rightarrow \mathcal{C}$, $\mathcal{R} \rightarrow \mathcal{P}$.
MiniDomainNet is a subset of DomainNet \cite{peng2019moment} containing 140K $96 \times 96$ images of 126 categories
from Clipart (Cl), Painting (Pa), Real (Re), and Sketch (Sk) domains.
We evaluate CadaFT on four MSDA tasks: $\mathcal{R} \rightarrow Cl$, $\mathcal{R} \rightarrow Pa$, $\mathcal{R} \rightarrow Re$ and $\mathcal{R} \rightarrow Sk$,
in which $\mathcal{R}$ means the remaining three domains apart from the target domain.

\textbf{Baseline Methods} TSA \cite{li2021transferable}, SFDA \cite{kim2021domain}, SHOT \cite{liang2020we}, AAA \cite{li2021divergence}, FAUST+U \cite{lee2023feature}, SCLM \cite{TANG2022467}, U-SFAN+ \cite{roy2022uncertainty}, AaD \cite{yang2022attracting}, C\&C \cite{chen2023contrast}, CoWA-JMDS \cite{lee2022confidence}, VMP \cite{lee2022confidence}, DIPE \cite{wang2022exploring}, ProxyMix \cite{ding2023proxymix}, Swin \cite{liu2022swin}, ViT \cite{dosovitskiy2020image}.
All baseline results are directly cited from the relevant published papers.

\textbf{Implementation details}. The foundation model $f \left( \bm{x}; \boldsymbol{\theta}^z \right)$ in CadaFT is implemented with ViT\footnote{Pre-trained weights from https://huggingface.co/timm/vit\_base\_patch16\_224.mae} and Swin\footnote{Pre-trained weights from https://huggingface.co/timm/swin\_base\_patch4\_window12\_384.ms\_in22k} respectively.
All compared models are trained on 4$\times$ V100 GPU by Adam optimizer \cite{jlb2015adam} with $\beta_1=0.9$,
$\beta_2=0.999$,
a low weight decay of $0.01$,
a mini-batch size $64$,
and the initial learning rate $1e-4$ with the linear warmup.

\begin{table}[htbp]
  \centering
  {\scriptsize
    \begin{tabular}{p{2.4cm} p{0.6cm} p{0.6cm} p{0.6cm} p{0.6cm} p{0.6cm} p{0.6cm} p{0.6cm} p{0.6cm} p{0.6cm} p{0.6cm} p{0.6cm} p{0.6cm} p{0.6cm}}
    \toprule
    Method & A$\rightarrow$C & A$\rightarrow$P & A$\rightarrow$R & C$\rightarrow$A & C$\rightarrow$P & C$\rightarrow$R & P$\rightarrow$A & P$\rightarrow$C & P$\rightarrow$R & R$\rightarrow$A & R$\rightarrow$C & R$\rightarrow$P & Avg. \\
    \midrule
    TSA \cite{li2021transferable}  & 57.6  & 75.8  & 80.7  & 64.3  & 76.3  & 75.1  & 66.7  & 55.7  & 81.2  & 75.7  & 61.9  & 83.8  & 71.2 \\
    SFDA \cite{kim2021domain} & 48.4  & 73.4  & 76.9  & 64.3  & 69.8  & 71.7  & 62.7  & 45.3  & 76.6  & 69.8  & 50.5  & 79.0  & 65.7 \\
    SHOT \cite{liang2020we} & 57.1  & 78.1  & 81.5  & 68.0  & 78.2  & 78.1  & 67.4  & 54.9  & 82.2  & 73.3  & 58.8  & 84.3  & 71.8 \\
    AAA \cite{li2021divergence}  & 56.7  & 78.3  & 82.1  & 66.4  & 78.5  & 79.4  & 67.6  & 53.5  & 81.6  & 74.5  & 58.4  & 84.1  & 71.8 \\
    FAUST+U \cite{lee2023feature} & 61.4  & 79.2  & 79.6  & 63.3  & 76.9  & 75.2  & 65.3  & 59.4  & 79.0  & 74.7  & 64.2  & 86.1  & 72.0 \\
    SCLM \cite{TANG2022467} & 58.2  & 80.3  & 81.5  & 69.3  & 79.0  & 80.7  & 69.0  & 56.8  & 82.7  & 74.7  & 60.6  & 85.0  & 73.1 \\
    U-SFAN+ \cite{roy2022uncertainty} & 57.8  & 77.8  & 81.6  & 67.9  & 77.3  & 79.2  & 67.2  & 54.7  & 81.2  & 73.3  & 60.3  & 83.9  & 71.9 \\
    AaD \cite{yang2022attracting}  & 59.3  & 79.3  & 82.1  & 68.9  & 79.8  & 79.5  & 67.2  & 57.4  & 83.1  & 72.1  & 58.5  & 85.4  & 72.7 \\
    C\&C \cite{chen2023contrast} & 59.0  & 79.5  & 82.0  & 67.6  & 79.2  & 79.5  & 66.7  & 56.5  & 81.3  & 74.2  & 58.3  & 84.7  & 72.4 \\
    CoWA-JMDS \cite{lee2022confidence} & 56.9  & 78.4  & 81.0  & 69.1  & 80.0  & 79.9  & 67.7  & 57.2  & 82.4  & 72.8  & 60.5  & 84.5  & 72.5 \\
    VMP \cite{lee2022confidence}  & 57.9  & 77.6  & 82.5  & 68.6  & 79.4  & 80.6  & 68.4  & 55.6  & 83.1  & 75.2  & 59.6  & 84.7  & 72.8 \\
    DIPE \cite{wang2022exploring} & 56.5  & 79.2  & 80.7  & 70.1  & 79.8  & 78.8  & 67.9  & 55.1  & 83.5  & 74.1  & 59.3  & 84.8  & 72.5 \\
    ProxyMix \cite{ding2023proxymix} & 59.3  & \textbf{81.0}  & 81.6  & 65.8  & 79.7  & 78.1  & 67.0  & 57.5  & 82.7  & 73.1  & 61.7  & 85.6  & 72.8 \\
    Swin \cite{liu2022swin} & 65.3  & 79.5  & 84.8  & 74.8  & 77.7  & 80.2  & 69.8  & 59.5  & 83.9  & 63.2  & 63.2  & 85.1  & 73.9 \\
    ViT \cite{dosovitskiy2020image}  & 56.2  & 74.6  & 78.7  & 65.2  & 72.9  & 77.6  & 65.8  & 52.3  & 78.9  & 73.7  & 58.2  & 83.6  & 69.8 \\
    \midrule
    CadaFT-Swin & 62.6  & 79.7  & 83.8  & 73.2  & 75.8  & 81.8  & 68.7  & 57.8  & 83.8  & 77.7  & 64.5  & 86.1  & 74.6 \\
    CadaFT-ViT & \textbf{66.1}  & 80.1  & \textbf{85.1}  & \textbf{76.8}  & \textbf{81.7}  & \textbf{85.3}  & \textbf{72.3}  & \textbf{66.8}  & \textbf{85.4}  & \textbf{79.8}  & \textbf{68.9}  & \textbf{86.8}  & \textbf{77.9} \\
    \bottomrule
    \end{tabular}}
    \caption{The averaged test accuracies over 5 runs on Office-home.}
    \label{tab:officehome}%
\end{table}%

\begin{table}[htbp]
  \centering
    \begin{tabular}{lccccc}
    \toprule
    Methods & $\mathcal{R} \rightarrow Cl$  & $\mathcal{R} \rightarrow Pa$  & $\mathcal{R} \rightarrow Re$  & $\mathcal{R} \rightarrow Sk$  & Avg. \\
    \midrule
    Source-only & 63.4  & 49.9  & 61.5  & 44.1  & 54.8 \\
    DANN \cite{ganin2016domain}  & 65.6  & 46.3  & 58.7  & 47.9  & 54.6 \\
    MCD \cite{saito2018maximum}   & 62.9  & 45.8  & 57.6  & 45.9  & 53.0 \\
    DCTN \cite{xu2018deep}  & 62.1  & 45.8  & 58.9  & 48.3  & 54.5 \\
    M$^3$SDA \cite{peng2019moment} & 64.2  & 49.1  & 57.7  & 49.2  & 55.0 \\
    MME \cite{saito2019semi}   & 68.1  & 47.1  & 63.3  & 43.5  & 55.5 \\
    DEAL \cite{zhou2021domain}  & 70.0  & 55.1  & 66.1  & 55.7  & 61.7 \\
    FAUST \cite{lee2023feature} & 68.1  & 52.2  & 68.7  & 59.1  & 62.0 \\
    FAUST+U \cite{lee2023feature} & 67.0  & 51.9  & 67.1  & 57.5  & 60.9 \\
    \midrule
    CadaFT-ViT & 71.4 & 59.3 & 68.1 & 52.6 & 62.9 \\
    CadaFT-Swin & \textbf{75.7} & \textbf{63.8} & \textbf{78.3} & \textbf{65.4} & \textbf{70.8} \\
    \bottomrule
    \end{tabular}%
    \caption{Classification accuracy (\%) on MiniDomainNet. The $\mathcal{R}$ denotes the remaining source domains}
    \label{tab:MiniDomainNet}%
\end{table}%

\textbf{Results Analysis.} The proposed CadaFT approach is evaluated against various baselines,
as shown in Tables \ref{tab:officehome} and \ref{tab:MiniDomainNet},
where CadaFT achieved SOTA domain adaptation results on both single-source and multi-source domain adaptations.
Specifically,
CadaFT-ViT achieved best results on 11 out of 12 SSDA tasks: A$\rightarrow$C, A$\rightarrow$R, A$\rightarrow$A, C$\rightarrow$P, C$\rightarrow$R, P$\rightarrow$A, P$\rightarrow$C, P$\rightarrow$R, R$\rightarrow$A, R$\rightarrow$C, R$\rightarrow$P with $66.1\%$, $85.1\%$, $76.8\%$, $81.7\%$, $85.3\%$, $72.3\%$, $66.8\%$, $85.4\%$, $79.8\%$, $68.9\%$, $86.8\%$ respectively.
CadaFT achieves a $2.0\%$ rise of averaged accuracy over all 12 SSDA tasks compared to previous baselines.
On Office-home,
CadaFT-Swin achieved the best results for all MSDA tasks with $75.7\%$ on $\mathcal{R}\rightarrow Cl$, $63.8\%$ on $\mathcal{R}\rightarrow Pa$, $78.3\%$ on $\mathcal{R}\rightarrow Re$ and $65.4\%$ on $\mathcal{R} \rightarrow Sk$ respectively,
and also the top-1 averaged accuracy of $70.8\%$.
Moreover,
comparing CadaFT variants and fined-tuned pre-trained vision models - ViT-small and Swin-small shows that the proposed approach significantly improves domain adaptation by using confounder controlling in an adversarial domain adaptation framework.

\section{Conclusion}
This study proposed a confounder balancing in adversarial domain adaptation for PLMs fine-tuning (CadaFT).
CadaFT tackles the confounder balancing method in ADA by introducing a minimax game with (1) the maximization step supervised by the task targets, domain indicators and confounders to attain a certain level of representation learning performance,
and (2) the minimization step continuously dilutes the discrimination between source and target domains.
This minimax game aims to improve the confounder-balanced and domain-invariant representation learning from foundation models,
e.g., PLMs.
Empirical results demonstrate that CadaFT achieves new SOTA OOD generalization in NLP and CV tasks.
Furthermore,
the plug-and-play confounder balancing module in CadaFT can improve the debiasing of spurious correlations compared to existing ADA baselines.

\section*{Acknowledgments}

This study is supported by grants from the National Key Research and Development Program of the Ministry of Science and Technology (Grant No. 2022ZD0116002), the Science and Technology Department of Guizhou Province (Grant No. Qiankehe Support[2022]General019), the National Social Science Foundation - Major Project (Grant No. 20ZD226), the National Key Research and Development Program of the Ministry of Science and Technology (Grant No. 2021ZD0113400), the Shenzhen Development and Reform Commission (Grant No. XMHT20190108009), the National Natural Science Foundation (Grant No. 62276075), the Guangdong Provincial Key Laboratory (Grant No. 2022B1212010005), the National Key Research and Development Program of China (Grant No. 2022ZD0115305, 2021ZD0112905), the Major Key Project of PCL (Grant N0. PCL2022D01) and the National Natural Science Foundation of China (Grant No. 62106115, 62006062 and 62176076), The Major Key Project of PCL (Grand No. PCL2021A06, PCL2022D01).

\bibliographystyle{unsrt}

\end{document}